# THPL: A Vision-to-Language Decision Support Framework for Rainbow Trout Feeding Management in RAS

Meng Liang[a,†], Guanbo Feng[a,†], Haozhuang Chi[b], Shilong Zhao[c], Zhixin Xiong[d], Yuhang He[a], Wenfeng Han[a], Tianhao Zhao[a], Zhihong Ma[a,*], Ying Liu[a,*]

**Affiliations**

[a] *College of Biosystems Engineering and Food Science, Zhejiang University, Hangzhou 310058, Zhejiang, PR China*

[b] *School of Mechanical and Aerospace Engineering, Nanyang Technological University, Singapore 639798, Singapore*

[c] *College of Marine Technology and Environment, Dalian Ocean University, Dalian 116023, Liaoning, PR China*

[d] *College of Engineering, Nanjing Agricultural University, Nanjing 210031, Jiangsu, PR China*

[†] Equal Contribution

[*]Corresponding Author： zhma_20@zju.edu.cn, liuyingzju@zju.edu.cn

**Abstract**

In Recirculating Aquaculture Systems (RAS), precision feeding serves as a critical prerequisite for minimizing operational costs and improving fish welfare. However, existing methodologies suffer from a lack of cognitive alignment between fish behavioral representations and aquaculture management knowledge, which impedes their direct translation into executable and interpretable feeding decisions. To address this challenge, this study proposes THPL, a generative feeding decision framework tailored for rainbow trout (*Oncorhynchus mykiss*) cultured in RAS. First, a Fishsort is utilized to extract schooling trajectories and establish an Activity Coefficient (AC) quantifying feeding intensity. Second, a Hierarchical Behavior Encoder (HBE) is constructed to sequentially model individual temporal progression and collective group dynamics using a Temporal Transformer and a Set Transformer, transforming trajectory tensors into a dual evidence behavioral representation comprising explicit physical tokens and implicit soft tokens. Finally, these behavioral tokens are integrated with masked multidimensional environmental parameters, aquaculture background data, and expert control rules to conduct Low Rank Adaptation (LoRA) fine-tuning on a large language model (LLM), followed by joint parameter optimization using counterfactual multimodal Direct Preference Optimization (mDPO) to reinforce causal reasoning capacity. Experimental results show that the AC exhibits a statistically significant monotonic positive correlation with expert annotated feeding intensity (Spearman $\rho = 0.925$, $p < 0.001$). Ablation experiments further indicate that the overall feeding decision accuracy improves from 33.33% for the text only baseline to 93.33% with the incorporation of dual evidence behavioral tokens. These findings confirm that continuous spatiotemporal behavioral tokens provide the necessary physical grounding for LLM, establishing precise alignment between fish spatiotemporal dynamics and expert feeding protocols. Compared with standard LoRA fine tuning, counterfactual mDPO alignment elevates overall decision accuracy from 93.33% to 96.67%, advances METEOR from 58.10% to 85.30%, reduces Self-BLEU-2 from 58.79% to 52.88%, and increases Distinct-3 from 6.68% to 7.81%, demonstrating that this strategy effectively suppresses templating and rectifies actuation biases while enriching adaptive reasoning diversity and reinforcing both causal consistency and operational safety. Overall, by integrating continuous spatiotemporal fish kinematics with causal reasoning in LLM, this study provides a novel decision support paradigm for precision feeding management in aquaculture.

## 1. Introduction

Precision feeding serves as a pivotal strategy for optimizing production efficiency within Recirculating Aquaculture Systems (RAS). Dynamically modulating feeding strategies according to fish feeding behavior can lower production costs while mitigating the adverse impacts of overfeeding on the aquaculture environment and fish welfare. In actual production, routine feeding decisions still predominantly rely on manual observation, empirical judgment, or predefined feeding schedules. Due to the lack of standardized, quantitative characterizations of schooling feeding states, feeding strategies struggle to adapt dynamically to behavioral variations, thereby escalating the risk of feed undersupply or oversupply. Feeding deviations lead to elevated feed conversion ratio (FCR), water quality deterioration, increased operating costs, and compromised fish welfare (Van Riel et al., 2023; Cho et al., 2024; Hvas et al., 2024). Therefore, there is a clear need to establish intelligent decision support systems capable of objectively quantifying fish schooling feeding states and dynamically adjusting feeding strategies accordingly, thereby elevating the precision, interpretability, and operational executability of feeding management.

In recent years, computer vision has been widely applied to fish feeding intensity assessment (FFIA) owing to its noninvasive nature, intuitive reflection of schooling behavior, and cost effectiveness. Building on this foundation, task-specific models (TSMs) based on machine learning and deep learning have further enhanced the accuracy and efficiency of feeding state quantification, providing quantitative criteria for feeding strategy optimization (Nielsen et al., 2016; Wang et al., 2021; Yang et al., 2021). In general, feeding intensity quantification methods can be categorized into indirect and direct approaches (Li et al., 2020). Indirect quantification mainly infers fish feeding states through secondary external manifestations such as surface disturbances or residual feed, including the analysis of light reflection, splash ripples, and the spatial distribution of uneaten feed. Recent deep learning methods have substantially improved the recognition performance of such manifestations, with splash feature based feeding intensity classification achieving accuracy exceeding 98.22% (Wu et al., 2024). However, although indirect visual indicators such as surface feeding splashes and uneaten pellets achieve high recognition accuracy in laboratory setups (Wu et al., 2024), discrepancies frequently persist between these secondary manifestations and authentic feeding behavior, impeding the stable and direct characterization of schooling feeding states (Li et al., 2020). Consequently, direct quantification methods that extract kinematic features such as sudden accelerations and sharp turns via TSMs have become an important pathway for objectively evaluating schooling feeding motivation (Fuchs and Caudill, 2019; Wei et al., 2021; Gyllingberg et al., 2023). Nevertheless, in high density RAS environments, visual occlusions, fluctuating target counts, and frequent identity switches readily induce trajectory fragmentation and noise, thereby constraining the stability and robustness of trajectory driven TSMs (Zhao et al., 2026). Furthermore, most TSM driven FFIA systems remain confined to behavioral quantification and state recognition, struggling to translate numerical perceptual outputs into practical management decisions. Although rule based methods can categorize feeding states according to predetermined thresholds, they lack a comprehensive understanding of intricate aquaculture contexts, failing to deliver operational recommendations that combine interpretability with executability (Zhou et al., 2018).

In comparison, Large Language Models (LLMs) demonstrate strong capabilities in contextual logical reasoning, multi-source information synthesis, and semantic consistency (Okutu and Yumetoshi, 2024). However, applying them to real-world aquaculture control imposes rigorous requirements on physical grounding, biological constraints, and operational safety. Without effective constraints, general LLMs are prone to factual hallucinations or outputs inconsistent with authentic production conditions when interpreting complex biological phenomena and generating management recommendations (Jiang et al., 2025; Narisetty et al., 2025). Furthermore, LLMs cannot directly interpret continuous video streams and fish schooling trajectories, creating a fundamental need to transform visual behavioral information into representations accessible to language models. Although prior studies have explored cross-modal alignment between visual or temporal sequences and language models via Querying Transformers (Li et al., 2023) or temporal prototype reprogramming (Jin et al., 2024), these methodologies predominantly target static visual features or regular one-dimensional time series. Consequently, they struggle to effectively characterize the complex collective dynamics caused by fluctuating target counts, visual occlusions, and identity switches in high-density aquaculture environments.

Specifically, when converting visual behavioral information into representations intelligible to language models, explicit kinematic features possess clear physical meanings and favorable interpretability, yet they struggle to fully preserve fine grained spatiotemporal information within intricate collective dynamics. In contrast, implicit soft tokens can encode richer dynamic behavioral patterns, but they lack directly verifiable physical grounding (Butt et al., 2026). Therefore, establishing a behavior to language interface that unifies explicit physical evidence and implicit soft tokens is essential to reconcile behavioral richness with physical interpretability while supporting reliable and executable feeding decisions. To address these challenges, this study proposes THPL, a knowledge grounded vision to language decision support framework tailored for rainbow trout (*O. mykiss*) cultured in RAS, aiming to systematically unify behavioral richness, physical interpretability, and operational safety. The primary contributions of this work are threefold. First, we construct a dimensionless Activity Coefficient (AC) integrating displacement, velocity, and acceleration, and design a hierarchical behavior encoder to jointly model explicit kinematic features and implicit behavioral tokens. Second, we construct a vision to language feeding decision architecture coupling counterfactual mDPO alignment with expert rule constraints, enhancing the diversity, safety, and reliability of decision generation. Third, we establish a comprehensive methodological pipeline bridging fish schooling behavioral representations to intelligent feeding decisions, providing technical support for precise, safe, and interpretable feeding management in complex aquaculture environments.

## 2. Materials and methods

This study constructed and validated the THPL decision support framework through a methodological workflow, bridging continuous quantitative behavioral perception with intelligent feeding reasoning in RAS. As schematically mapped across the following subsections, the complete experimental and methodological pipeline comprises four interconnected phases. First, Section 2.1 details the controlled RAS experimental setup, video acquisition protocols, feeding trial design for *O. mykiss*, and the physical-consistent kinematic augmentation strategy used to construct the benchmark dataset. Second, Section 2.2 delineates the four-tiered system architecture, systematically covering the TSM for trajectory tracking and AC calculation, the HBE for extracting dual-evidence behavioral tokens, the PET for compiling structured environmental and rule prompts, and the active LLM decision engine. Third, Section 2.3 presents the multi-stage training pipeline, which sequentially coordinates HBE feature pretraining, multimodal supervised fine-tuning, and counterfactual mDPO alignment to reinforce physical causality and prevent decision trivialization. Finally, Section 2.4 outlines a multi-dimensional evaluation protocol to rigorously verify rule execution accuracy, operational safety boundaries, linguistic generation diversity, and causal decision stability against sensory perturbations.

### *2.1. Experimental design and dataset construction*

#### *2.1.1 Experiment system and data acquisition*

This experiment was conducted as a controlled pilot study at Yuhang Smart Aquaculture Research Center (Hangzhou, Zhejiang Province, China). Building upon the Fishsort framework established by Zhao et al. (2026), the present study conducts further investigations into collective behavioral modeling, with specific experimental procedures and technical parameters regarding fish detection and multi-target tracking thoroughly documented in their original publication. The experimental setup is shown in Fig. 1. A total of 78 *O. mykiss* with an average body weight of 65g were placed in a RAS tank (1.60 m × 1.60 m). Throughout the experiment, the water temperature in the tank was maintained at 12 °C, pH level was maintained within the optimal range of 8.0, and the dissolved oxygen concentration was consistently maintained above 8 mg/L. The nutritional balance of the experimental feed was ensured by using commercially extruded feed specifically formulated for *O. mykiss*. Prior to the commencement of the experiment, the fish were subjected to a 7-day acclimatisation period in the rearing tanks. During the temporary rearing period and the 10-day formal experiment, the fish were fed at fixed times of 8:30 and 16:30 every day. Feeding commenced at 5-second intervals until the fish reached a satiated state and ceased feeding. Ten minutes after feeding concluded, remaining feed was removed to prevent water contamination.

The experimental data acquisition system comprised three main components, including video recording, water quality monitoring, and data processing units. A camera was mounted 1 m above the water surface to capture top-view footage, with a frame rate of 20 FPS and a resolution of 1280 × 720. Video data from each feeding session were stored in a digital video recorder (DS-7816N-16P, Hikvision, China). The camera system started recording 2 minutes before feeding began and stopped recording when fish no longer feed, capture various feeding behavior states of fish school. During the experiment, the changes in the intensity of *O. mykiss* feeding were observed and recorded. In the water quality monitoring unit, a handheld sensor (DO200, YSI, USA) was used to measure dissolved oxygen (DO) and temperature during feeding.

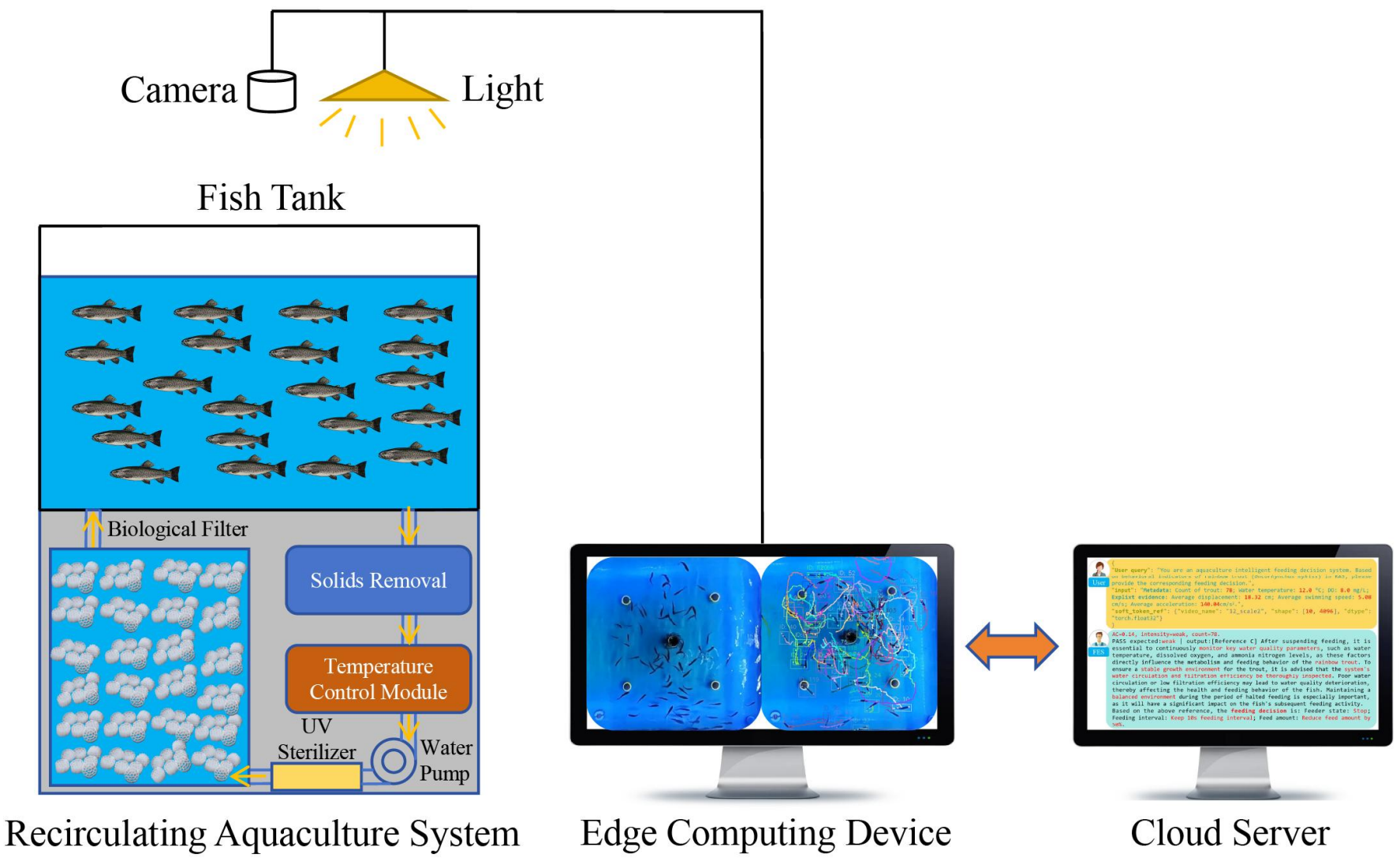

**Fig. 1.** Fish feeding monitoring, computing and analyzing in Recirculating Aquaculture Systems.

### *2.1.2. Data preprocessing, annotation, and Augmentation*

A total of 299 five second video segments were isolated chronologically across complete feeding trials conducted over ten consecutive days, with each day comprising two discrete feeding sessions conducted in the morning and afternoon respectively. All acquired video recordings were converted into consecutive image frames of 720 × 720 pixels. Ground truth feeding intensity categories $y^{\text{expert}} \in$ {Weak, Medium, Strong} were directly adopted from the benchmark dataset established in Zhao et al. (2026), where segments were evaluated by aquaculture experts following established behavioral criteria adapted from Wei et al. (2021). The dataset was partitioned chronologically into a training set of 239 raw segments from Days 1 to 8 (16 sessions), a validation set of 30 segments from Day 9 (2 sessions), and an independent test set of 30 segments from Day 10 (2 sessions) at an approximate 8 to 1 to 1 ratio. Because all recordings originated from a single continuous *O. mykiss* cohort, the 30 evaluation segments from Day 10 represent discrete behavioral evaluation units sampled across two separate feeding events following metabolic reset, evaluating within cohort temporal generalization rather than cross population biological replicates. Within Day 10, the 30 test segments were uniformly sampled across the two sessions to provide 10 balanced instances for each feeding intensity regime.

To enhance model generalization without introducing unphysical artifacts, a Physical Consistent Kinematic Augmentation (PCKA) pipeline was applied strictly to the 239 raw training segments, generating 2,868 transformed kinematic views to construct a composite training corpus of 3,107 instances (detailed in Table A.1, **Appendix A**). Because transformations primarily relied on rigid spatial rotations and reflections that preserve scalar velocity magnitudes, augmented instances naturally inherited their parent feeding intensity labels. The validation and test sets (30 raw samples each) remained strictly unaugmented raw sequences to ensure unbiased performance evaluation. Kinematic parameters were strictly recalculated from transformed spatial coordinates post augmentation using finite differences to maintain internal kinematic coherence. Bivariate correlation scatter matrices and two sample Kolmogorov Smirnov tests indicated that these geometric transformations did not induce severe statistical distributional drift or extreme numerical artifacts in low order kinematics ($p > 0.05$), though these empirical operations do not imply full hydrodynamic equivalence (**Appendix A.1**, Figs. A.1 and A.2). The resulting trajectory tensors and derived kinematic features were subsequently formatted into standardized sequence representations to serve as structured inputs for downstream behavioral modeling and decision reasoning.

### *2.2. Overall architecture of behavior-aware decision system*

#### *2.2.1. System architecture*

As schematically illustrated in Fig. 2, the THPL framework operates as an integrated decision architecture comprising four cohesive functional tiers, including the task-specific behavioral perception tier (TSM), the hierarchical behavioral encoding tier (HBE), the prompt engineering tier (PET), and the policy decision tier (LLM). Within the perception tier, Fishsort executes multi-object tracking across top-view feeding video streams to extract continuous coordinates and construct trajectory sequences. Simultaneously, kinematic preprocessing extracts individual displacement, swimming velocity, and acceleration to compute the macroscopic Activity Coefficient (AC). Calibrated by data driven optimal thresholds, this metric reclassifies collective feeding vigor into weak, medium, and strong regimes, redirecting behavioral states to index expert rules within the PET and guide auxiliary classification supervision in HBE. Building upon these continuous trajectories, the HBE transforms trajectory coordinates into spatiotemporal tensors and cascades an entity-level Temporal Encoder with a group-level Set Encoder via a Temporal-Set Bidirectional Cross-Attention module to capture multi-agent dynamics without tracking order dependency. Under auxiliary class loss supervision, an explicit behavioral encode head pools macroscopic physical metrics into explicit tokens, while an implicit behavioral encode head distills high-dimensional dynamics into implicit soft tokens, which are concatenated into continuous behavioral prefix embeddings. In parallel, the PET coordinates a rule engine and prompt composer to formulate domain knowledge, organizing structured numerical context including water temperature and dissolved oxygen, categorical metadata, and expert knowledge rules indexed by the supervisory AC range into structured text tokens alongside standardized task queries. At the decision tier, the active policy model (Llama-3.1-8B with trainable LoRA) receives both the concatenated behavioral prefix tokens and structured text tokens, aligned through a counterfactual mDPO loss engine against a frozen reference model to reinforce causal reasoning. During online inference, the active model performs behavior-grounded reasoning to generate structured decision text encompassing feed amount, feeding interval, feeder state, and expert rationale, which is strictly verified by a deterministic output validator before dispatching actuation commands to automated feeders.

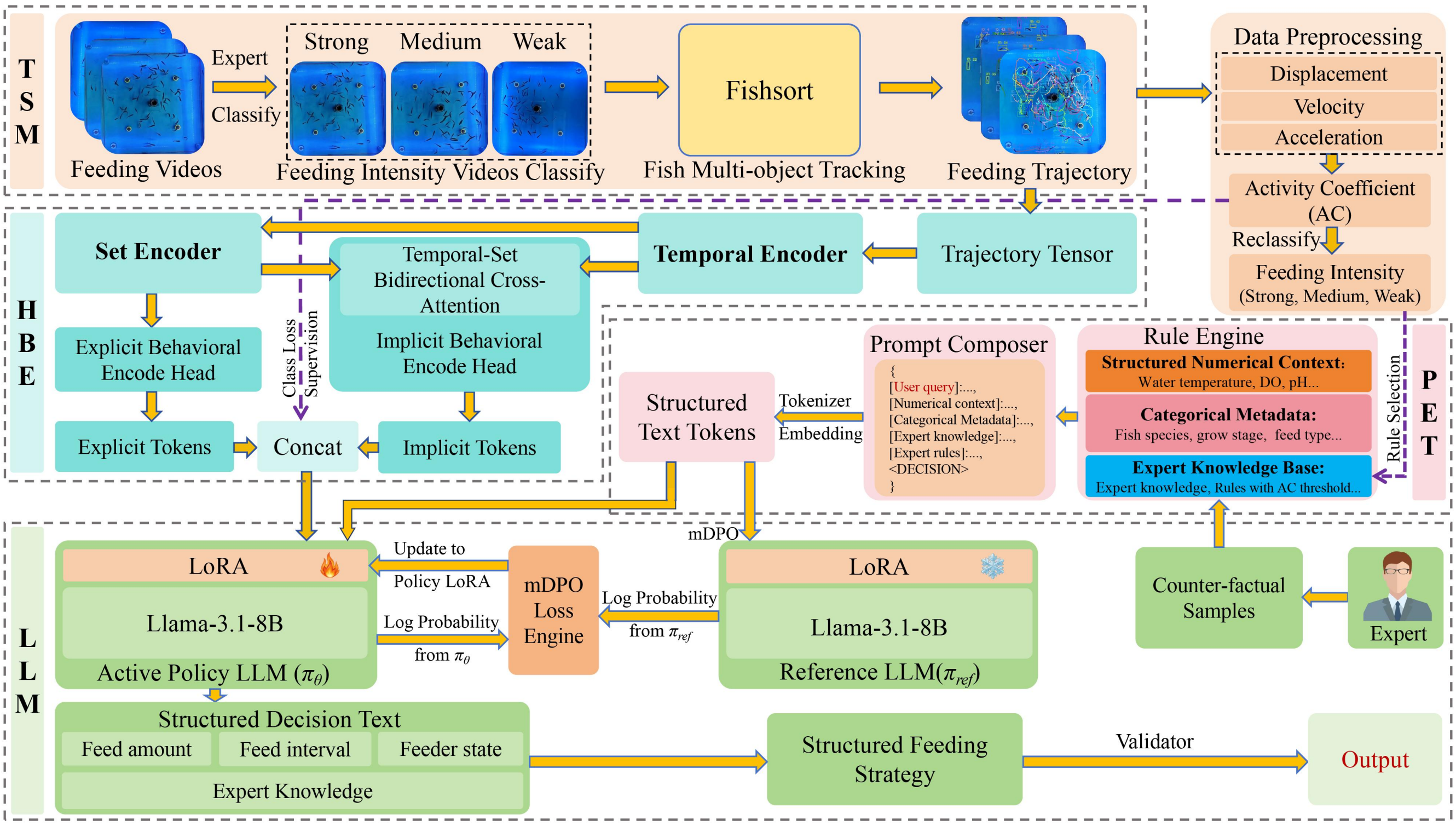

**Fig. 2.** Architectural overview of the THPL framework. The task-specific model (TSM) perception tier extracts multi-fish trajectories and computes Activity Coefficients (AC) to index candidate rules. The hierarchical behavior encoder (HBE) tier combines temporal and set encoders with bidirectional cross-attention to produce concatenated explicit and implicit behavioral tokens. The prompt engineering tier (PET) compiles environmental telemetry and expert rules into structured text tokens. The large language model (LLM) decision tier, fine-tuned via Low-Rank Adaptation (LoRA) and aligned through counterfactual multimodal Direct Preference Optimization (mDPO), synthesizes structured feeding actions verified by a deterministic post-generation validator.

*2.2.2. Fish school trajectory tracking and Activity Coefficient formulation via TSM*

In the TSM, Fishsort was employed to track fish schools across video sequences (Zhao et al., 2026), extracting continuous trajectories to construct a spatiotemporal trajectory tensor $X \in \mathbb{R}^{B \times T \times N \times F}$ across T temporal segments ($T$=10) and $N$ tracked fish. To mitigate high frequency tracking oscillations and surface turbulence, each five second video sequence was uniformly discretized into 10 temporal windows of 0.5 s. As aligned with the encoder input space, this trajectory tensor encapsulates five explicit kinematic channels (F=5) comprising normalized centroid coordinates (cx, cy), planar velocity components ($vx$, $vy$), and instantaneous acceleration ($acc$). Concurrently, key kinematic parameters were computed across individuals via finite differences to formulate the macroscopic Activity Coefficient (AC). Individual displacement ($D_{i,t}$) measures cumulative path length within each segment to reflect spatial foraging coverage. Swimming velocity ($V_{i,t}$) captures instantaneous planar speed, and acceleration ($A_{i,t}$) reflects burst turning dynamics. Because $D$ and $V$ covary across duration, an equal-weighting fusion was adopted after individual z-score standardization to balance cumulative exploration and burst kinetics, supported by the ablation in **Appendix A**. Variables $\kappa \in \{D, V, A\}$ were normalized using the global mean ( $\mu_\kappa^{\text{train}}$ ) and standard deviation ( $\sigma_\kappa^{\text{train}}$ ) derived strictly from the 239 raw training sequences. As formulated in Eq. (1), the raw indicator $AC_t^{\text{raw}}$ averages these standardized features across valid fish within segment $t$. Finally, Eq. (2) scales this raw coefficient into a strictly bounded continuous interval using the global extrema from the training set to facilitate threshold partitioning, where $\epsilon$ constitutes a small constant preventing division by zero.

$$AC_t^{\text{raw}} = \frac{1}{3} \sum_{\kappa \in \{D,V,A\}} \left( \frac{1}{n_t} \sum_{i=1}^{n_t} \frac{\kappa_{i,t} - \mu_\kappa^{\text{train}}}{\sigma_\kappa^{\text{train}} + \epsilon} \right) \tag{1}$$

$$AC_t = \min\left(1, \max\left(0, \frac{AC_t^{\text{raw}} - AC_{\min^{\text{raw,train}}}}{AC_{\max^{\text{raw,train}}} - AC_{\min^{\text{raw,train}}} + \epsilon}\right)\right) \tag{2}$$

In addition to the mean-based formulation in Eqs. (1) and (2), candidate aggregation formulations based on group variance and kinetic energy were designed to evaluate physical representation efficacy across different statistical moments (detailed in **Appendix A.2**). To partition the continuous AC into discrete feeding intensity regimes, a data-driven grid-search optimization protocol was executed strictly on the independent validation set to determine optimal decision boundaries while preventing data leakage (formally defined in **Appendix A.3**, Eqs. A.4-A.6). Through this optimal thresholding procedure, the scalar AC reclassifies collective feeding intensity into weak, medium, and strong regimes as illustrated in Fig. 2. This reclassification establishes a dual redirection mechanism within the architecture. It redirects discrete behavioral states to index corresponding candidate control rules within PET, while concurrently supplying supervisory targets for auxiliary classification losses during HBE pretraining. Furthermore, to verify whether the derived kinematic index captures authentic biological phase transitions rather than optical tracking artifacts, local fish school density was incorporated as a spatial metric to evaluate its biomechanical alignment with collective feeding dispersal (detailed in **Appendix A.4**).

### *2.2.3. Hierarchical Behavior Encoder*

As illustrated in Fig. 3, the HBE maps variable-length tracking data into fixed-size embedding tokens compatible with the LLM embedding space. To preserve localized kinematics while handling dynamic target quantities, the HBE employs a cascaded Temporal-then-Set structure driven by attention mechanisms (**Appendix B**, Eqs. B.1 to B.4). The input trajectory tensor $\mathbf{X}\in\mathbb{R}^{B\times T\times N\times 5}$ is augmented with sinusoidal temporal position encodings along dimension $T$. A key-padding mask filters out invalid coordinates and untracked padding instances across frames. Operating across the temporal dimension $T$ for each tracked fish independently, an individual-level Temporal Transformer leverages multi-head self-attention (Vaswani et al., 2017) to capture swimming dynamics and burst turning kinetics. This encoder extracts entity-level representations $\mathbf{H}_{temp}\in\mathbb{R}^{B\times N\times Dt}$ from the terminal valid temporal step. Subsequently, a Set Transformer processes $\mathbf{H}_{temp}$ across the entity dimension $N$ via Set Attention Blocks (Lee et al., 2019). This generates unpooled interaction tokens $\mathbf{H}_{set}\in\mathbb{R}^{B\times N\times Ds}$ that preserve permutation equivariance over fish track indices without bottleneck compression.

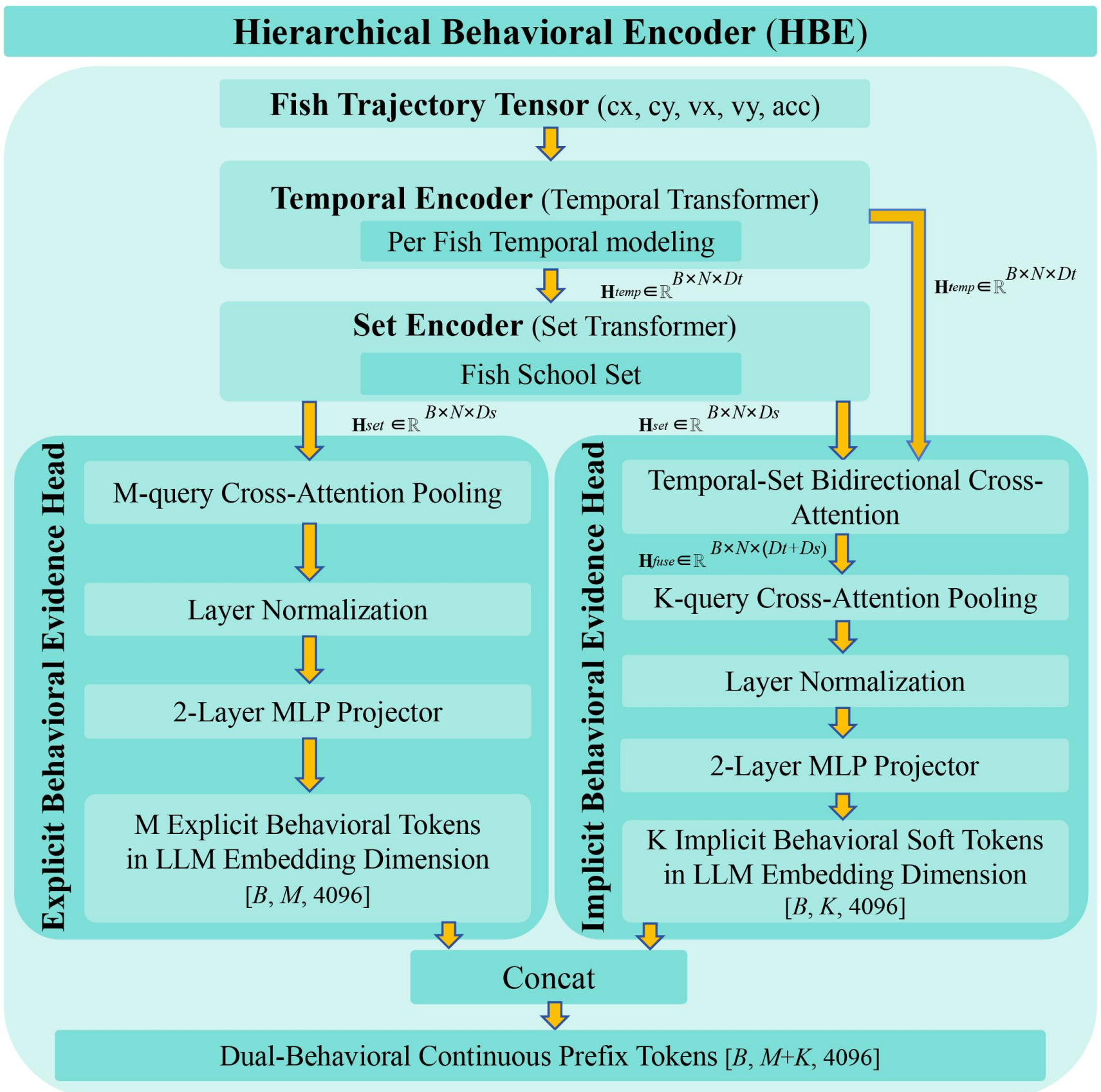

**Fig. 3.** Architecture of the Hierarchical Behavior Encoder. Trajectory tensors are processed through cascaded temporal and set Transformers to capture individual burst kinematics and permutation-invariant collective dynamics. An explicit head extracts macroscopic physical tokens, while an implicit head

applies temporal-set bidirectional cross-attention to distill fine-grained soft tokens. Both token streams are concatenated into continuous prefix embeddings for language model grounding.

Building upon these representations, an explicit evidence head models macroscopic physical motion cues across displacement, velocity, acceleration, and school density. It applies $M$ learnable queries $\mathbf{Q}_M \in \mathbb{R}^{B\times M\times D_s}$ to cross-attention pool $\mathbf{H}_{set}$, yielding explicit behavioral tokens $Z_{exp} \in \mathbb{R}^{B\times M\times 4096}$ as defined in Eq. (3). To ensure dimensional compatibility with LLM while maintaining gradient propagation, these explicit tokens are projected into the continuous embedding space rather than unnormalized physical scalars, acting as physically targeted inductive priors. In parallel, an implicit behavioral head fuses individual kinematics $\mathbf{H}_{temp}$ with collective context $\mathbf{H}_{set}$ via Temporal-Set Bidirectional Cross-Attention. Specifically, it computes temporal-contextualized features $\mathbf{H}_{t\to s} = \text{Multihead}(\mathbf{H}_{temp}, \mathbf{H}_{set}, \mathbf{H}_{set})$ and set-contextualized features $\mathbf{H}_{s\to t} = \text{Multihead}(\mathbf{H}_{set}, \mathbf{H}_{temp}, \mathbf{H}_{temp})$. These features are concatenated along the channel dimension to construct the joint behavioral context $\mathbf{H}_{fuse} = [\mathbf{H}_{t\to s}; \mathbf{H}_{s\to t}] \in \mathbb{R}^{B\times N\times (D_t+D_s)}$ Then, $K$ learnable queries $\mathbf{Q}_K \in \mathbb{R}^{B\times K\times (D_t+D_s)}$ pool high-dimensional dynamics from $\mathbf{H}_{fuse}$ to generate implicit behavioral soft tokens $\mathbf{Z}_{soft} \in \mathbb{R}^{B\times K\times 4096}$ via Eq. (4). Through query-based attention pooling, both $Z_{exp}$ and $Z_{soft}$ achieve strict permutation invariance with respect to global fish indexing order. Rather than entering the encoder as a direct numeric input, the scalar AC functions strictly as an offline supervisory criterion to partition feeding regimes for multi-task loss supervision. The explicit and soft tokens are concatenated along the sequence dimension into dual-behavioral prefix embeddings $Z_{out} \in \mathbb{R}^{B\times (M+K)\times 4096}$, propagating gradients to ground downstream linguistic reasoning in visual kinematics.

$$Z_{exp} = \text{MLP}_{exp}\left\{\text{LayerNorm}\left[\text{Multihead}(\mathbf{Q}_M, \mathbf{H}_{set}, \mathbf{H}_{set})\right]\right\} \tag{3}$$

$$Z_{soft} = \text{MLP}_{soft}\left\{\text{LayerNorm}\left[\text{Multihead}(\mathbf{Q}_K, \mathbf{H}_{fuse}, \mathbf{H}_{fuse})\right]\right\} \tag{4}$$

### *2.2.4. Rule Engine and Prompt Composer*

While the HBE generates continuous behavioral embeddings, the rule engine and prompt composer construct the semantic textual context to activate the reasoning capabilities of LLM without updating rule weights. As summarized in Table 1, the expert rule catalog defines operational control directives indexed by offline AC thresholds. The discrete parameters (5 s, 7 s, 10 s intervals; 20%, 50% reductions) serve as pre-configured heuristic baselines. For TH001 (Weak), the feeder halts immediately, with the 10 s interval defining a standby re-observation latency and the 50% quota specifying a provisional restart probing dose. To prevent shortcut retrieval, runtime prompts embed candidate rule specifications but exclude numerical AC values and ground-truth labels. By pairing the 239 raw training trajectories with sampled water quality conditions (dissolved oxygen, temperature, and ammonia), 4,500 structured interaction records were synthesized. Among them, 3,000 records were allocated to Stage 2 supervised fine-tuning, while the remaining 1,500 paired records provided counterfactual rationales for Stage 3 preference alignment. Crucially, the independent validation split ($N$=30) remained completely isolated from this synthesis pipeline, sharing no parent trajectories, prompt variations, or counterfactual pairs with the training corpus.

**Table 1**

Expert rule catalog and offline behavioral intensity partitioning criteria based on AC.

| Rule ID | Feeding Intensity | AC Condition | Feeder State | Feeding Interval | Feed Amount |
|---|---|---|---|---|---|
| TH003 | Strong | $0.462 < AC \le 1.00$ | Continue | Keep 5s | Keep 100% |
| TH002 | Medium | $0.181 < AC \le 0.462$ | Continue | Keep 7s | Reduce by 20% |
| TH001 | Weak | $0.00 \le AC \le 0.181$ | Stop | Keep 10s | Reduce by 50% |

Note: Ground-truth feeding intensity is established via expert annotation (Section 2.1.2). AC ranges serve strictly as offline thresholds for rule indexing and loss supervision. Under TH001 (Weak), the feeder halts immediately. The 10 s interval defines a standby re-observation latency, and the 50% quota specifies a provisional restart probing dose. Operational parameters (5 s, 7 s, 10 s; 20%, 50%) are pre-configured expert heuristic baselines. Inference runs directly on behavioral prefix tokens without AC inputs.

The prompt composer dynamically formats heterogeneous inputs into an instruction prompt $\mathcal{P}$ by concatenating discrete contextual blocks, as expressed in Eq. (5). Specifically, $\mathcal{C}_{num}$ supplies qualitative water quality descriptors such as temperature, pH and dissolved oxygen (DO) condition levels, $\mathcal{C}_{meta}$ specifies aquaculture background parameters including species, biomass and tank specifications, $\mathcal{C}_{rule}$ injects candidate expert rules encompassing operational boundaries across 3 feeding regimes, $\mathcal{Q}_{user}$ provides the standardized task query, and $\langle \text{DECISION} \rangle$ acts as the generation delimiter. Prepending the behavioral prefix tokens to the embedded textual prompt establishes a unified input sequence at the embedding layer, compelling the model to ground decisions in visual kinematics while respecting environmental constraints. Formally, the sequential composition of the instruction prompt $\mathcal{P}$ is defined in Eq. (5).

$$\mathcal{P} = \mathcal{C}_{num} \,\|\, \mathcal{C}_{meta} \,\|\, \mathcal{C}_{rule} \,\|\, \mathcal{Q}_{user} \,\|\, \langle \text{DECISION} \rangle \tag{5}$$

During parameter-efficient fine-tuning, the prompt text functions strictly as an unpenalized conditioning context rather than an autoregressive prediction target. A standard label mask of -100 is assigned across all positions corresponding to both the prepended behavioral tokens and the instruction prompt $\mathcal{P}$, excluding them from cross-entropy loss computation. Gradients are propagated exclusively over the subsequent response tokens containing structured control commands and expert diagnostic rationales, ensuring the model acquires domain reasoning without memorizing prompt templates.

### 2.2.5. *Parameter-efficient LoRA fine-tuning, counterfactual mDPO alignment and validator*

As illustrated in Fig. 2, the active policy language model is adapted and aligned through parameter-efficient fine-tuning coupled with mDPO (Wang et al., 2024) driven by counterfactual expert pairs. In this workflow, LoRA (Hu et al., 2021) is introduced by freezing the pre-trained weights $W_0$ and training low-rank decomposition matrices ($r$=8, $\alpha$=16) across attention projection layers (**Appendix B**, Eq. B.5). To resolve unconditional preference shortcuts where multimodal architectures bypass visual inputs, the policy is aligned via counterfactual mDPO without training an external reward network. As shown in the mDPO optimization loop of Fig. 4, expert-curated counterfactual preference pairs are fed into both the active policy $\pi_\theta$ and a frozen reference policy $\pi_{ref}$. The mDPO loss engine computes log-probability ratios between $\pi_\theta$ and $\pi_{ref}$ conditioned on behavioral prefix tokens to optimize the active LoRA weights, directly parameterizing the implicit reward function formulated in Eq. (6):

$$r(v,\mathcal{P},y)=\beta\log\frac{\pi_\theta(y\mid v,\mathcal{P})}{\pi_{ref}(y\mid v,\mathcal{P})} \tag{6}$$

where $v$ denotes behavioral prefix tokens ($Z_{out}$), y denotes candidate decision text, and $\beta$ governs the deviation temperature. To reinforce causal consistency across the environment-behavior-decision chain, counterfactual preference pairs were curated across five key operational dimensions comprising water temperature, dissolved oxygen, stocking density, feeding history, and cross-species physiological baselines (detailed in Table C.1, **Appendix C**). Crucially, counterfactual perturbations hold swimming kinematics $v$ strictly invariant while altering environmental conditions to acute stress regimes, defining restrictive actuator actions as chosen responses $y_w$ and unconstrained feeding continuation as rejected responses $y_l$. During inference, a deterministic output validator parses the generated text, verifying predicted feeder states, feeding intervals, and feed amounts against predefined expert thresholds before execution.

## 2.3. *Multi-Stage Training Strategy*

The unified training pipeline of the proposed architecture coordinates hierarchical optimization across three sequential phases to ensure representation fidelity, multimodal instruction following, and causal decision grounding, schematically illustrated in Fig. 4.

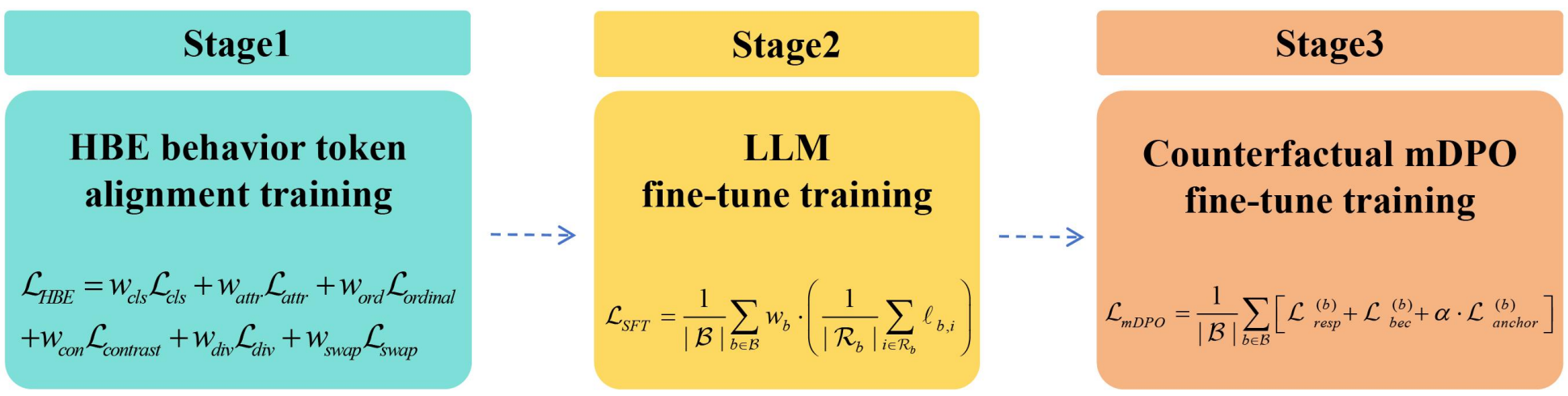

**Fig. 4.** Schematic overview of the three-stage training pipeline encompassing HBE behavior token alignment pretraining (Stage 1), LLM with LoRA training

(Stage 2), and counterfactual mDPO with LoRA training (Stage 3).

In Stage 1, the HBE is supervised by a composite multi-task loss function to extract discriminative and physically grounded features. Specifically, the core classification loss $\mathcal{L}_{cls}$ acts as the primary task supervisor across a mini-batch of size $B$ (indexed by sample $b$), computed via cross-entropy on the mean-pooled features $h_b$ of both explicit evidence tokens and soft tokens to distinguish feeding intensity according to Eq. (7). To explicitly preserve continuous physical properties, the attribute supervision loss $\mathcal{L}_{attr}$ formulated in Eq. (8) forces the pooled soft representations $\mathbf{s}_b = \text{MeanPool}(\mathbf{Z}_{soft,b}) \in \mathbb{R}^{4096}$ of each sample $b$ to regress true kinematic indicators via an attribute prediction head $\text{MLP}_{attr}$ against ground-truth kinematic targets $y_b$. An ordinal classification loss $\mathcal{L}_{ordinal}$ defined in Eq. (9) enforces the hierarchical relationship across the three ordered feeding levels ($C$=3) using $R = C - 1 = 2$ cumulative binary classifiers $f_r$. To bridge the representation gap between physical metrics and latent features, the contrastive alignment loss $\mathcal{L}_{contrast}$ expressed in Eq. (10) enforces cross-branch consensus between pooled soft representations $\mathbf{s}_b$ and explicit representations $\mathbf{e}_b = \text{MeanPool}(\mathbf{Z}_{exp,b}) \in \mathbb{R}^{4096}$ across the batch. To prevent token dimensional degeneration and feature redundancy, a token diversity loss $\mathcal{L}_{div}$ detailed in Eq. (11) penalizes excessive cosine similarity between distinct soft tokens among the $K$ queries beyond a margin threshold $\delta$. Furthermore, a structural sensitivity loss $\mathcal{L}_{swap}$ presented in Eq. (12) enforces temporal sensitivity with a margin $m$ by penalizing the network if pooled soft representations remain invariant when temporal frame orders are randomly shuffled ( $\widetilde{\mathbf{X}}$ ). The final composite HBE pre-training objective $\mathcal{L}_{HBE}$ established in Eq. (13) integrates these individual loss components through a weighted summation, where the weighting coefficients $[w_{cls}, w_{attr}, w_{ord}, w_{con}, w_{div}, w_{swap}]$ are empirically configured to [1.0, 1.0, 1.0, 0.5, 0.1, 0.1] respectively.

$$\mathcal{L}_{cls} = -\frac{1}{B}\sum_{i=1}^{B}\log\frac{\exp(z_{i,y_i})}{\sum_{c=1}^{C}\exp(z_{i,c})} \tag{7}$$

$$\mathcal{L}_{attr} = \frac{1}{B}\sum_{b=1}^{B}\| \text{MLP}_{attr}(s_b) - y_b \|_2^2 \tag{8}$$

$$\mathcal{L}_{ordinal} = -\frac{1}{B}\sum_{b=1}^{B}\sum_{r=1}^{R}\left[\mathbb{I}(y_b \ge r)\log\sigma\big(f_r(\mathbf{s}_b)\big) + \mathbb{I}(y_b < r)\log\big(1-\sigma(f_r(\mathbf{s}_b))\big)\right] \tag{9}$$

$$\mathcal{L}_{contrast} = -\frac{1}{B}\sum_{b=1}^{B}\log\frac{\exp\left(\mathrm{sim}(\mathbf{s}_b,\mathbf{e}_b)/\tau\right)}{\sum_{j=1}^{B}\exp\left(\mathrm{sim}(\mathbf{s}_b,\mathbf{e}_j)/\tau\right)} \tag{10}$$

$$\mathcal{L}_{div} = \frac{1}{K(K-1)}\sum_{i=1}^{K}\sum_{j\neq i}^{K}\max(0,\mathrm{sim}(s_i,s_j)-\delta) \tag{11}$$

$$\mathcal{L}_{swap} = \frac{1}{B}\sum_{b=1}^{B}\max\left(0, m - \| \mathbf{s}_b(\mathbf{X}) - \mathbf{s}_b(\widetilde{\mathbf{X}})\|_2^2\right) \tag{12}$$

$$\mathcal{L}_{HBE} = w_{cls}\mathcal{L}_{cls} + w_{attr}\mathcal{L}_{attr} + w_{ord}\mathcal{L}_{ordinal} + w_{con}\mathcal{L}_{contrast} + w_{div}\mathcal{L}_{div} + w_{swap}\mathcal{L}_{swap} \tag{13}$$

In Stage 2, the LLM policy $\pi_\theta$ undergoes supervised fine-tuning to map the combined behavioral prefix tokens and structured prompt text into structured textual decisions. The *K* soft tokens $s_b \in \mathbb{R}^{K\times H}$ and explicit tokens are prepended directly to the textual prefix as continuous prompt embeddings, where the fundamental single-token optimization target $\ell_{b,i}$ follows standard causal language modeling formulated in Eq. (14). To mitigate classification boundary drift associated with sparse minority events, a weighted Supervised Fine-Tuning loss $\mathcal{L}_{SFT}$ formulated in Eq. (15) is applied across the batch $\mathcal{B}$. This loss is computed exclusively over the valid response tokens $\mathcal{R}_b$ by assigning an ignore index of -100 to all prefix and prompt positions. While this masking restricts loss computation to response generation rather than prompt auto-completion, trainable parameters remain conditioned on the context. Mitigating rigid response templating is primarily governed by downstream counterfactual mDPO alignment. Weak intensity samples are upweighted ( $w_{weak} = 2.0$ ) to emphasize the critical penalty of feeding during anomalous states, while other classes retain a standard weight ( $w_b = 1.0$ ).

$$\ell_{b,i} = -\log p_\theta(y_{b,i} \mid y_{b,<i}, s_b) \tag{14}$$

$$\mathcal{L}_{SFT} = \frac{1}{|\mathcal{B}|}\sum_{b\in\mathcal{B}} w_b \cdot \left(\frac{1}{|\mathcal{R}_b|}\sum_{i\in\mathcal{R}_b}\ell_{b,i}\right) \tag{15}$$

To guarantee feature space consistency prior to Stage 3, behavioral prefix tokens were extracted offline using the strictly frozen, final checkpoint of the trained HBE and projector from Stage 2. The cached embedding files were version-locked and verified via cryptographic checksums against the corresponding model weights to eliminate latent feature drift across training stages. In Stage 3, the policy $\pi_\theta$ undergoes preference optimization adapted from the mDPO framework to enforce conditional reliance on trajectory evidence and resolve unconditional language shortcuts. This optimization objective adapts three log-sigmoid terms to evaluate the trainable policy $\pi_\theta$ against a frozen reference model $\pi_{ref}$. The response preference term $\mathcal{L}_{resp}$ formulated in Eq. (16) compels the model to favor the verified decision $y_w$ over the counterfactual error $y_l$. When environmental safety thresholds enforce feeding termination, paired responses $y_w$ and $y_l$ differ in kinematic diagnostic rationales rather than forcing an invalid action flip. The behavior-evidence conditional preference term $\mathcal{L}_{bec}$ detailed in Eq. (17) contrasts response probabilities given valid trajectory tokens $v$ against an unconditioned baseline $\varnothing$ (zero-vector masking), penalizing the policy if it relies solely on language priors while ignoring physical inputs. The anchored preference term $\mathcal{L}_{anchor}$ expressed in Eq. (18) regularizes training by penalizing absolute likelihood degradation on chosen responses relative to $\pi_{ref}$. Finally, the batch-level objective $\mathcal{L}_{mDPO}$ in Eq. (19) aggregates these components using deviation temperature $\beta = 0.1$ and anchor weight $\alpha = 1.0$.

$$\mathcal{L}_{resp} = -\log\sigma\left[\beta\Big((\log\pi_\theta(y_w) - \log\pi_\theta(y_l)) - (\log\pi_{ref}(y_w) - \log\pi_{ref}(y_l))\Big)\right] \tag{16}$$

$$\mathcal{L}_{bec} = -\log\sigma\left[\beta\begin{pmatrix}(\log\pi_\theta(y_w \mid v) - \log\pi_\theta(y_w \mid \varnothing)) - \\ (\log\pi_{ref}(y_w \mid v) - \log\pi_{ref}(y_w \mid \varnothing))\end{pmatrix}\right] \tag{17}$$

$$\mathcal{L}_{anchor} = -\log\sigma\left[\beta(\log\pi_\theta(y_w) - \log\pi_{ref}(y_w))\right] \tag{18}$$

$$\mathcal{L}_{mDPO} = \frac{1}{|\mathcal{B}|}\sum_{b\in\mathcal{B}}\left[\mathcal{L}_{resp}^{(b)} + \mathcal{L}_{bec}^{(b)} + \alpha\cdot\mathcal{L}_{anchor}^{(b)}\right] \tag{19}$$

*2.4. Evaluation of LLM-generated feeding decisions*

To comprehensively assess the performance of the fine-tuned large language model in the aquaculture feeding domain, evaluations were conducted on a balanced test set comprising 30 independent video segments (10 samples for each of the weak, medium, and strong feeding intensity classes). The evaluation framework is bifurcated into rule-constrained instruction execution and natural language semantic generation. The rule-constrained execution evaluation assesses the ability of the model to generate safe and actionable control commands conforming to expert protocols. Distinct from visual annotations ($y^{\text{expert}}$) and auxiliary kinematic supervision ($y^{\text{AC}}$), target labels in this stage denote expert reference policy actions $\mathbf{a}^{\text{policy}} = (s^*, t^*, a^*)$ specified by standardized aquaculture management regimes. Discrete action match rates evaluate individual prediction accuracy across feeder state, feeding interval, and feed amount. Because expert feeding regimes define a bijective mapping between schooling appetite and the tripartite actuator actions, overall decision accuracy ($\text{Acc}_{dec}$) requires simultaneous exact concordance across all three fields as formulated in Eq. (20), rendering overall accuracy bounded by the lowest performing dimension and quantifying execution compliance against prescriptive expert rules. To strictly penalize catastrophic errors such as issuing feeding commands during abnormal weak appetite, the Unsafe Continue Rate (USCR) is calculated via Eq. (21). Additional classification metrics including per-class recall and invalid generation rate are documented in **Appendix B** (Eqs. B.6-B.8).

$$\mathrm{Acc}_{dec} = \frac{1}{N}\sum_{i=1}^{N}\prod_{j}\mathbb{I}(\hat{y}_{i,j} = y_{i,j}) = \frac{1}{N}\sum_{i=1}^{N}\mathbb{I}(\hat{\mathbf{y}}_i = \mathbf{y}_i) \tag{20}$$

$$USCR = \frac{N_{unsafe}}{N_{total}} \times 100\% \tag{21}$$

The natural language semantic evaluation examines the linguistic fidelity, factual consistency, and reasoning robustness of generated rationales. Semantic similarity and cognitive diversity metrics encompassing BLEU, Self-BLEU, ROUGE-L, METEOR, and BERTScore are detailed in **Appendix B** (Eqs. B.9-B.16). To eliminate factual fabrications in water quality parameters or control logic, the Hallucination Penalty Score (HPS) formulated in Eq. (22) deducts severity-weighted penalties from a base score $S_{base}$ according to detected factual discrepancies $I(h_i)$. Furthermore, to evaluate causal robustness against tracking noise, the decision consistency rate $\mathcal{C}(\delta)$ defined in Eq. (23) measures the exact retention of generated action triplets under trajectory perturbation level $\delta$ relative to unperturbed executions.

$$HPS = S_{base} - \lambda\sum_{i=1}^{K}\omega_i \cdot I(h_i) \tag{22}$$

$$\mathcal{C}(\delta) = \frac{1}{N}\sum_{i=1}^{N}\mathbb{I}\left(\hat{\mathbf{y}}_i^{(\delta)} = \hat{\mathbf{y}}_i^{\mathrm{clean}}\right) \times 100\% \tag{23}$$

## 3. Results

### *3.1. Kinematic modeling, threshold optimization, and spatial distribution verification*

The quantitative characterization of fish school kinetics was established through the composite AC, which integrates individual displacement, velocity, and acceleration into a macroscopic indicator. To determine the optimal mathematical formulation for physical behavior aggregation among the candidate formulations designed in Section 2.2.2, a comparative evaluation investigating the standardized mean ($AC_{mean}$), variance ($AC_{var}$), and magnitude energy ($AC_{energy}$) was conducted across the experimental dataset. One-way ANOVA and effect size analyses confirmed that the mean-based formulation ($AC_{mean}$) provided the highest between-group discriminatory capability, achieving an ANOVA F-statistic of 176.88 ($p < 0.001$) and an effect size $\eta^2$ of 0.9291, markedly outperforming the variance and energy formulations. The complete mathematical definitions, detailed statistical distribution comparisons, pairwise Cohen's d effect sizes, and correlation scatter matrices supporting this empirical selection are comprehensively documented in **Appendix A** (Figs. A.3-A.6). The performance degradation observed in the energy-based formulation occurs because zero-centered standardized deviations fold negative values into positive quadratic magnitudes, whereas physical kinetic energy models require non-negative raw velocity fields. Consequently, $AC_{mean}$ was definitively selected as the optimal kinematic aggregation metric.

As depicted in Fig. 5(a), the continuous AC exhibits a rigorous monotonic concordance with ground-truth feeding intensity rankings (Spearman $\rho = 0.925$, $p < 0.001$; Kendall $\tau = 0.804$), substantiated by a Jonckheere-Terpstra trend test ($z = 17.79$, $p < 0.001$). In Fig. 5(b), the kernel density distributions of the selected index across the 299 benchmark sequences display distinct behavioral boundaries corresponding to the three feeding intensity classes. Implementing the grid-search optimization protocol established in Section 2.2.2 yielded optimal decision boundaries at [0.00, 0.181] for weak feeding, (0.181, 0.462] for medium feeding, and (0.462, 1.00] for strong feeding. Under this optimal thresholding strategy, the AC derived from fish school trajectory features achieves a baseline classification accuracy of 95.32% (macro-F1 = 95.43%), correctly identifying 114 out of 117 weak feeding samples, 107 out of 113 medium feeding samples, and 64 out of 69 strong feeding samples. Furthermore, a stratified five-fold cross-validation protocol with fold-independent parameter re-estimation across the 299 benchmark sequences yielded an out-of-fold generalization accuracy of 94.31% (282/299 samples) under stable thresholds ($t_w = 0.181 \pm 0.003$, $t_m = 0.462 \pm 0.004$), confirming that the learned boundaries reflect authentic biomechanical phase transitions rather than sample-specific overfitting. The statistical consistency of the underlying kinematic data was further validated prior to model training, where bivariate correlation matrices and Kolmogorov-Smirnov tests confirmed that the augmented kinematic distributions exhibited no significant discrepancy from the raw experimental trajectories with all p-values exceeding 0.05, as comprehensively detailed in **Appendix A** (Figs. A.1 and A.2).

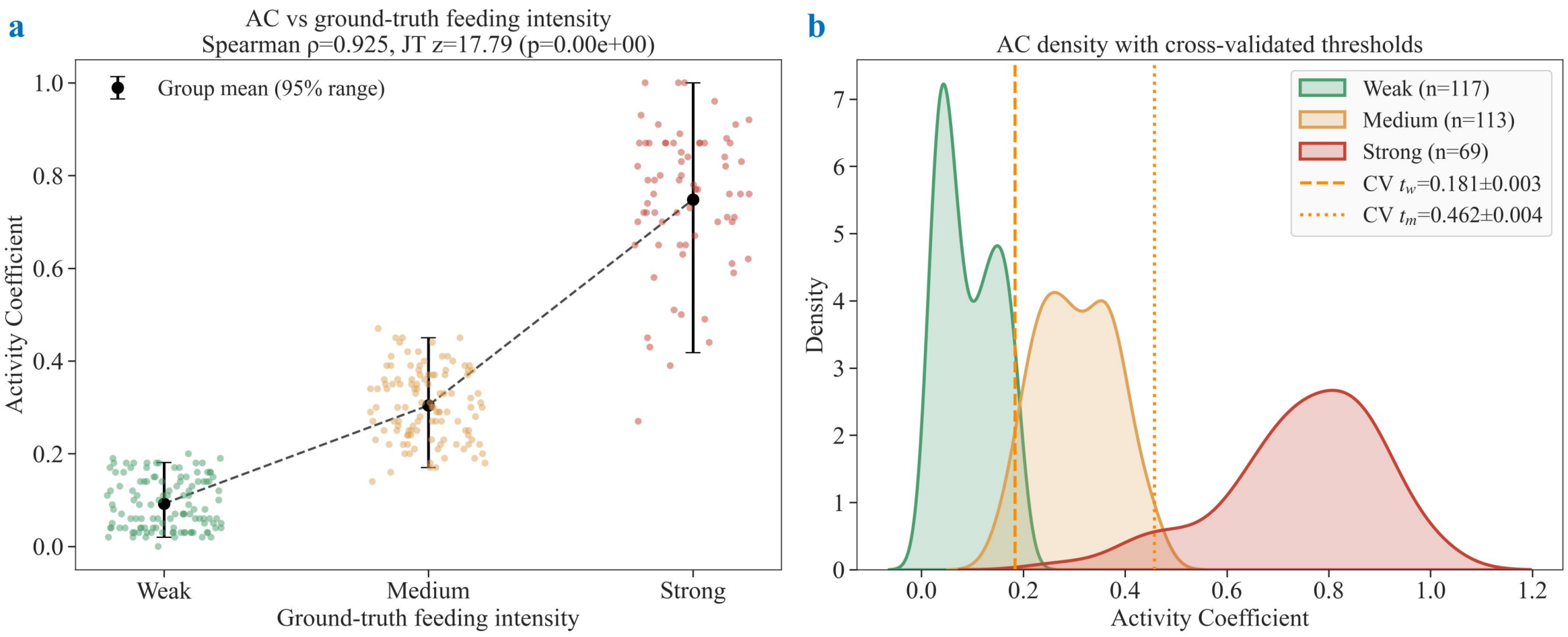

**Fig. 5.** Evaluation of the Activity Coefficient (AC) across 299 benchmark sequences. (a) monotonic correlation and trend verification against ground-truth feeding intensities with group means, 95% confidence intervals, and colored scatter distributions (Spearman $\rho = 0.925$, $p < 0.001$). (b) class-wise kernel

density distributions alongside cross-validated data-driven decision boundaries ($t_w$ = 0.181± 0.003, $t_m$ = 0.462± 0.004).

Corroborating the fidelity of the AC against spatial schooling distributions across 299 evaluation sequences confirmed that the quantified feeding intensity reflects physical schooling patterns. As documented in **Appendix A** (Fig. A.7), statistical evaluation demonstrated an exceptionally high negative correlation between local fish density and the AC ($r$ = -0.9629, $p$ < 0.001). The bivariate joint density heatmap in Fig. A7(a) reveals a continuous, high-concentration density progression from explosive foraging states toward dense resting aggregation, supported by a highly significant Chi-square association across behavioral regimes ($\chi^2$= 426.82, $p$ < 0.001, Fig. A7(b)). This empirical concordance confirms that the derived behavioral prior reliably captures the macro-level transition from cohesive aggregation during resting periods to spatial dispersal during voracious feeding.

*3.2. Architectural ablation, disturbance robustness, and representation probing of HBE*

To evaluate the internal mechanism of the HBE, systematic ablation studies were conducted using a five-fold cross-validation protocol. As illustrated in Fig. 6, the spatiotemporal topology comparison demonstrates that the proposed Temporal-then-Set architecture achieved the superior performance, attaining an accuracy of 97.0% and an F1-score of 0.967. Configurations lacking independent temporal sequence modeling for individual targets suffered substantial performance collapse, where the Set-Only and Set-then-Temporal models achieved accuracies of only 33.5% and 33.4% with F1-scores of 0.168 and 0.166, respectively. Although the Temporal-Only model achieved an accuracy of 90.2% and an F1-score of 0.895, it remained inferior to the proposed structure. These ablation outcomes confirm that accurately characterizing fish feeding states requires hierarchical modeling, wherein resolving individual temporal locomotion must precede collective aggregation to preserve transient burst kinematics while capturing permutation invariant group dynamics.

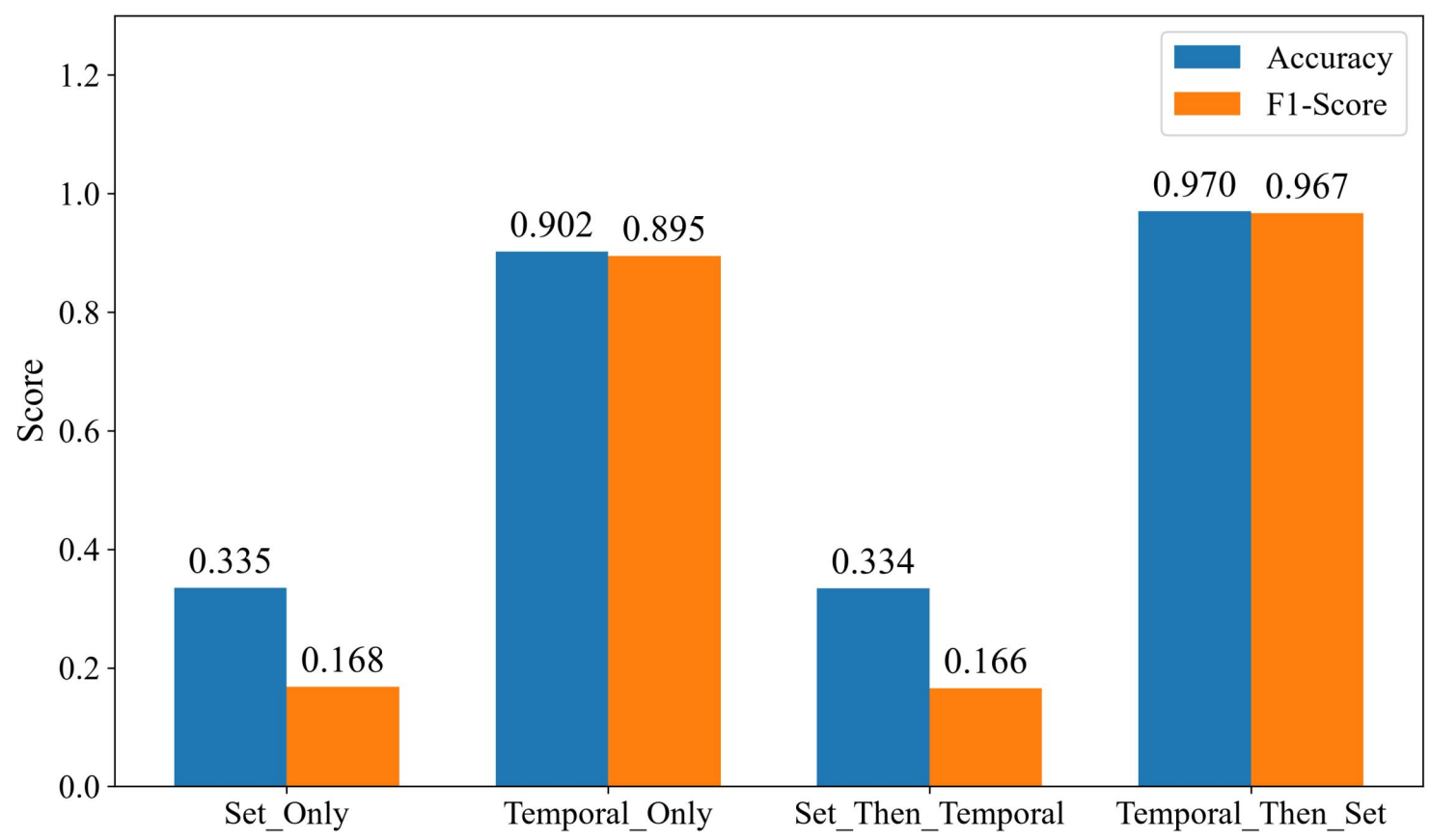

**Fig. 6.** Architectural ablation evaluations of the HBE including structural comparisons and input feature ablation under five-fold cross-validation.

Perturbation stress tests confirm that the policy network performs causal reasoning over continuous kinematics rather than relying on text. As shown in Fig. 7(a), corrupting trajectory inputs induces a monotonic decline in decision consistency rate relative to unperturbed executions, measuring the exact joint retention of feeder state, feeding interval, and feed amount. Minor perturbations at Tier 1 maintain 100.0% consistency, demonstrating an effective noise tolerance margin against subtle tracking oscillations. Increasing spatial Gaussian jitter steadily reduces consistency to 66.7% at the maximum severity tier, while temporal frame drop preserves 90.0% consistency under 70% frame loss. Completely ablating trajectory inputs with zeros collapses decision consistency to 33.3%, which matches the text-only baseline and confirms that structured action synthesis causally requires physical motion inputs. Kinematic distributions in Fig. 7(b) establish feeding states of these spatial features. Mean swimming velocity increases from 0.045 m/s in weak feeding to 0.121 m/s in strong feeding with a positive Spearman correlation of 0.95, whereas local spatial density drops from 542.4 to 321.5 with an inverse correlation of -0.93 as fish disperse during competitive feeding. These statistical patterns confirm that continuous trajectory coordinates preserve authentic biological dynamics, allowing the model to synthesize control decisions directly from continuous motion feature.

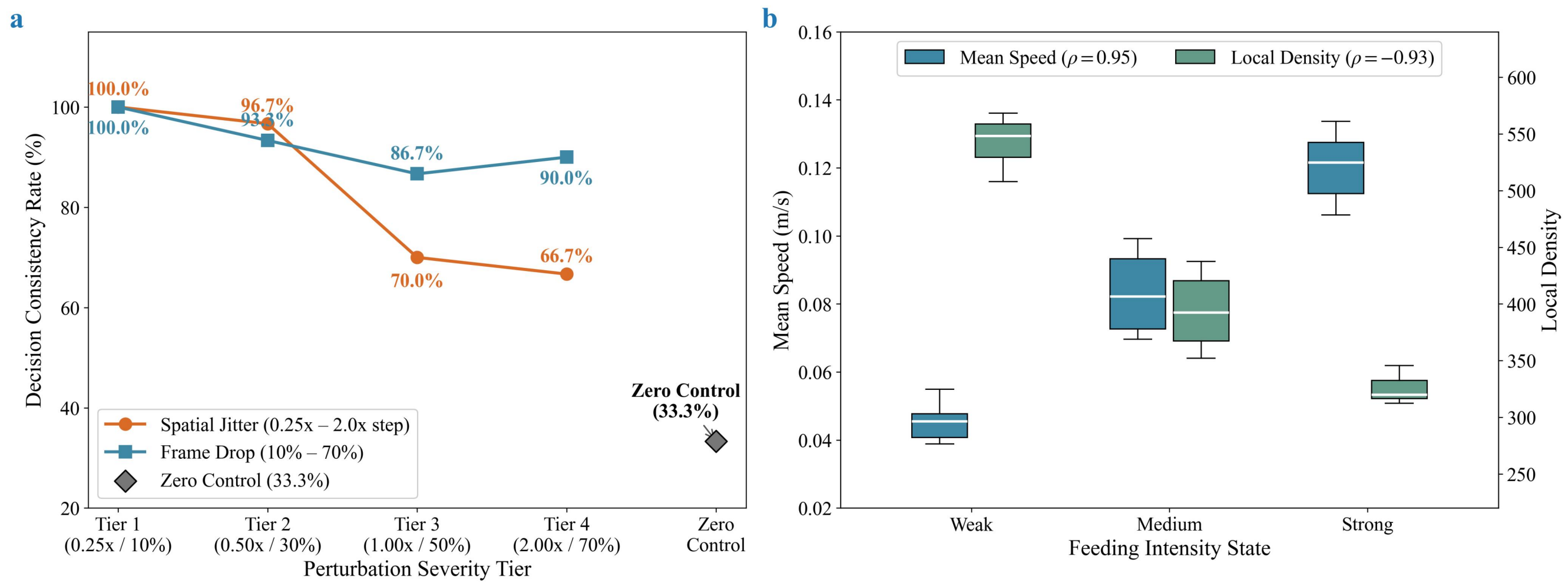

**Fig. 7.** Trajectory validity and distributions of spatial feature in HBE and LLM, showing (a) decision consistency rate across perturbation severity tiers under spatial jitter, temporal frame drop, and zero-trajectory control, where consistency denotes exact joint matching across feeder state, feeding interval, and feed amount relative to unperturbed inputs, and (b) distributions of mean swimming velocity and local spatial density across feeding intensity states.

In addition to continuous tracking noise, discrete identity switches caused by frequent fish occlusions represent another primary disturbance in dense aquaculture. To assess HBE robustness against tracking errors in dense aquaculture, artificial identity switches were introduced at 50, 100, 150, and 200 events per sequence. As shown in Fig. 8, the downstream LLM decision performance maintained complete stability with a variation delta of +0.0% across all swap frequencies. Specifically, Fig. 8(a) shows that the feeding intensity classification evaluated from the LLM's generated decisions preserved an overall accuracy of 93.3%, yielding 100.0% for weak, 100.0% for medium, and 80.0% for strong feeding under all conditions. In Fig. 8(b), the three structured decision action fields generated by the active policy LLM maintained identical match rates of 100.0% for feeder state, 93.3% for feeding interval, and 93.3% for feed amount. These results confirm that the temporal-then-set aggregation in HBE effectively decouples collective behavioral dynamics from individual tracking order, ensuring perturbation-free LLM reasoning.

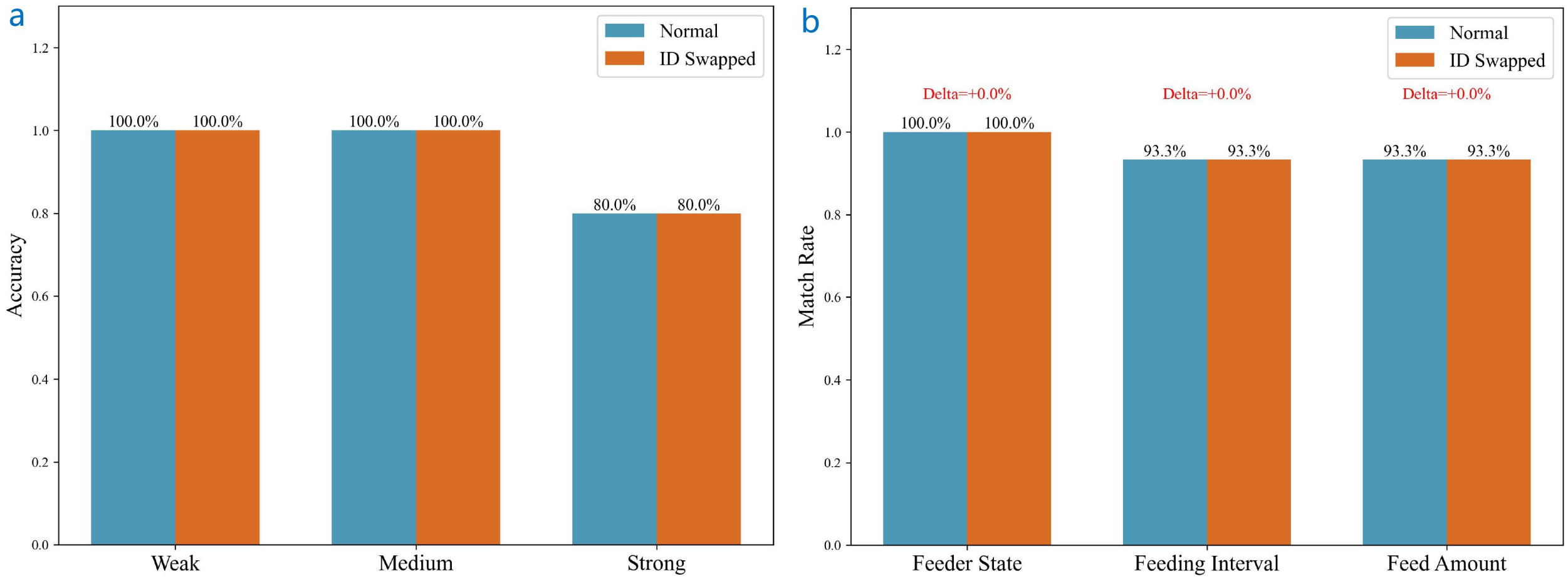

**Fig. 8.** Robustness of downstream LLM (SFT) decision-making under 50, 100, 150, and 200 tracking ID switches, showing (a) LLM-derived classification accuracy across feeding intensities and (b) match rates of generated decision action fields.

Alongside collective spatial dispersion, fine-grained kinematic analyses revealed an intrinsic behavioral heterogeneity that cannot be captured by scalar activity metrics. As documented in **Appendix A** (Fig. A.9), sequences possessing comparable macroscopic AC baselines consistently bifurcate into distinct behavioral geometries, comprising heterogeneous distributions characterized by an actively foraging minority alongside an inactive majority, as well as homogeneous distributions reflecting uniform cruising across the entire school. To verify whether the Hierarchical Behavior Encoder preserves these underlying distribution geometries without dimensional information bottlenecks, linear regression probing under five-fold cross-validation was conducted across the extracted behavioral tokens. As visualized in Fig. A.8 (**Appendix A**), the learned continuous tokens demonstrate remarkable fidelity in recovering both the microscopic velocity dispersion quantified by the coefficient of variation ($R^2$ = 0.851, $r$ = 0.923, $p$ < 0.001) and the macroscopic AC ($R^2$ = 0.894, $r$ = 0.951, $p$ < 0.001). In sharp contrast, explicit kinematic features alone completely fail to recover velocity dispersion ($R^2$ = -0.035). These empirical outcomes substantiate that the hierarchical continuous tokens successfully preserve higher-order velocity distribution moments, mitigating distributional insensitivity and equipping the downstream language model with robust physical representations.

### *3.3. Dual-evidence bottleneck verification*

To investigate the individual and synergistic contributions of behavioral token representations, ablation experiments were conducted across four input configurations comprising the text only baseline with 0 tokens, the explicit only configuration with M=4, the soft only configuration with K=10, and the full dual evidence configuration with M+K=14. As illustrated in Fig. 9(a), overall decision accuracy ($Acc_{dec}$) scales positively with the progressive incorporation of behavioral embeddings. The text only baseline yields an overall accuracy of only 33.33%, bounded by the lowest match rate across actuator fields. The confusion matrix in Fig. 9(c) reveals that this baseline suffers from severe majority class trivialization, predicting the medium feeding category for all thirty evaluation instances across weak, medium, and strong states due to the total absence of visual motion cues. Incorporating explicit physical tokens elevates overall decision accuracy to 46.67%, while utilizing implicit soft tokens alone achieves an accuracy of 86.67%. The full dual evidence model secures the highest decision accuracy of 93.33%, maintaining clear classification boundaries with only two strong feeding instances conservatively classified as medium feeding as depicted in Fig. 9(d).

The field specific evaluation presented in Fig. 9(b) highlights how dual evidence representations govern multi-dimensional control execution. In feeder state regulation, the soft only and full dual evidence configurations each achieve 100.00% match rates, outperforming the 66.67% rate obtained by explicit tokens alone and the text only baseline. For quantitative parameters, the complementary properties of the two feature streams become prominent. The feeding interval match rate reaches 93.33% under the full configuration, compared to 46.67% for explicit only, 86.67% for soft only tokens, and 33.33% for the text only baseline. Similarly, the feed amount match rate attains 93.33% in the full configuration, markedly improving upon the 46.67% achieved by explicit tokens alone, 86.67% achieved by soft tokens alone, and 66.67% from the text only baseline. Crucially, the dual-evidence synergy between explicit and implicit behavioral tokens drives the average decision accuracy across feeder state, feeding interval, and feed amount to 95.56%, substantially surpassing 91.11% obtained with implicit tokens alone, 53.33% with explicit tokens alone, and 55.56% from the text only baseline. These empirical outcomes confirm that continuous kinematic embeddings provide indispensable physical representation anchors for the language model, effectively mitigating decision bias in the absence of visual perception and establishing precise causal alignment between collective spatiotemporal dynamics and expert control protocols.

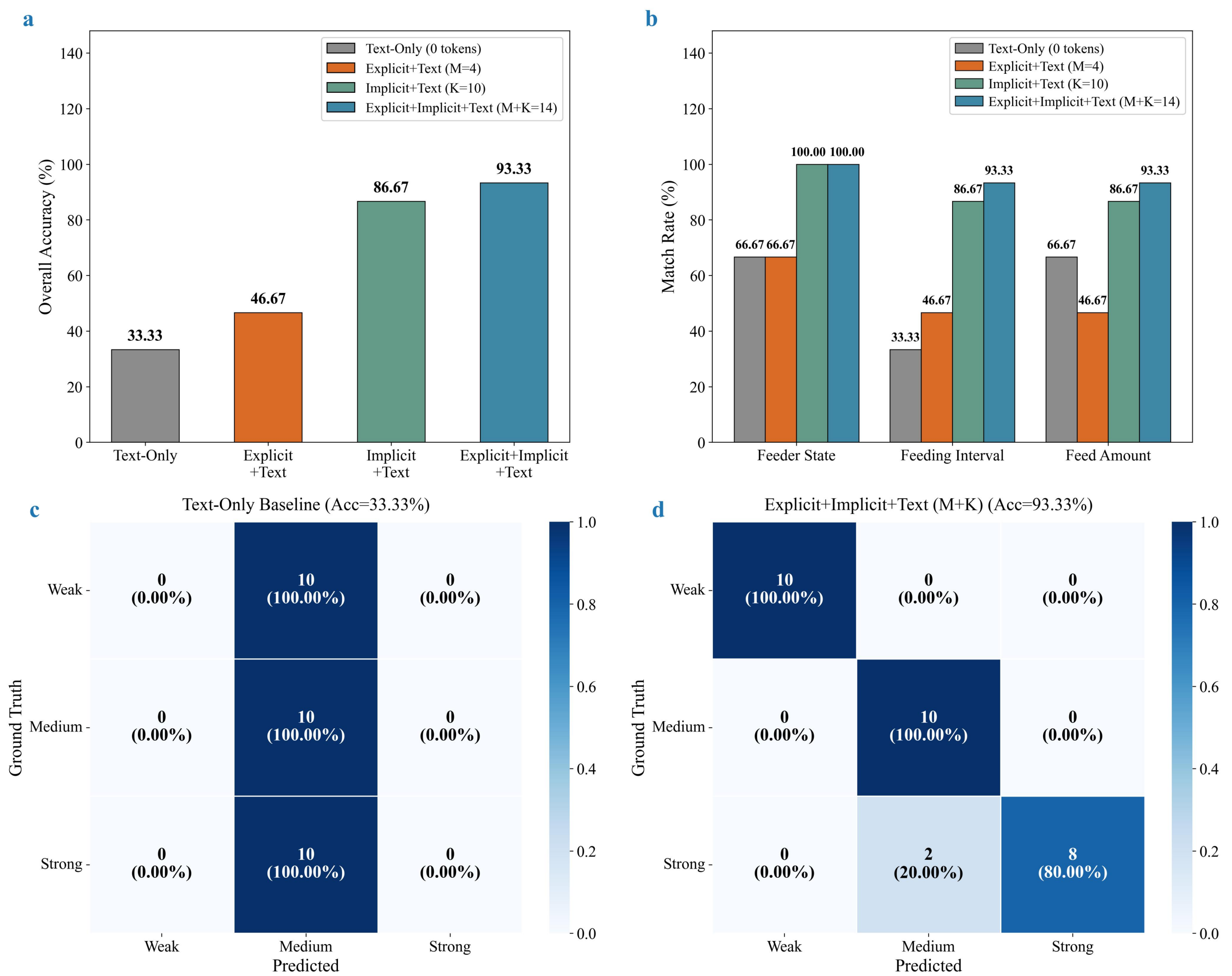

**Fig. 9.** Ablation evaluation of behavioral token configurations across decision performance and error distributions, showing (a) overall feeding decision accuracy ($Acc_{dec}$) across four input representations, (b) field-level match rates across three structured decision actions, (c) confusion matrix for the text only baseline demonstrating complete majority class collapse, and (d) confusion matrix for the full dual-evidence model.

### *3.4. LLM Structured decision performance, safety, and rule compliance*

The decision-making capacity of the fine-tuned LLM was evaluated across thirty independent test samples representing the three distinct feeding intensities. As summarized in Table 2, counterfactual mDPO alignment substantially improved operational decision fidelity over standard Supervised Fine-Tuning (LoRA). The overall decision accuracy increased from 93.33% to 96.67%, with the macro recall elevating symmetrically to 96.67% while preserving a 0.00% invalid rate across all testing instances. In the baseline LoRA configuration, the model exhibited underfitting on intensive foraging states, where strong feeding yielded an accuracy of only 80.00% due to a conservative bias that misclassified 20.00% of strong instances as medium feeding. Implementing counterfactual mDPO completely resolved this classification bottleneck, elevating strong feeding accuracy to 100.00% while maintaining 100.00% accuracy in weak feeding identification. Across the three discrete control fields, counterfactual mDPO achieved matching rates of 96.67% for feeder state, feeding interval, and feed amount, where the sole discrepancy was a benign conservative adjustment from medium to weak feeding, confirming high execution precision in discrete actuator regulation.

**Table 2**

Structured decision performance detailing per-class accuracy, recall, and field-wise match rates across feeding intensities.

| Optimization Strategy | Feeding Intensity | Accuracy (%) | Recall (%) | Match Rate (%) | | |
|---|---|---|---|---|---|---|
| | | | | Feeder State | Feeding Interval | Feed Amount |
| Supervised Fine-Tuning (LoRA) | Weak | 100.00 | 100.00 | 100.00 | 100.00 | 100.00 |
| | Medium | 100.00 | 100.00 | 100.00 | 100.00 | 100.00 |
| | Strong | 80.00 | 80.00 | 100.00 | 80.00 | 80.00 |
| | Average | 93.33 | 93.33 | 100.00 | 93.33 | 93.33 |
| Counterfactual mDPO + LoRA | Weak | 100.00 | 100.00 | 100.00 | 100.00 | 100.00 |
| | Medium | 90.00 | 90.00 | 90.00 | 90.00 | 90.00 |
| | Strong | 100.00 | 100.00 | 100.00 | 100.00 | 100.00 |
| | Average | 96.67 | 96.67 | 96.67 | 96.67 | 96.67 |

Note: Evaluations were conducted across 30 independent test video sequences, comprising 10 balanced samples per intensity class.

The operational safety and adherence to expert protocols were quantified under deterministic boundary validation. Both the baseline LoRA and preference-aligned mDPO models achieved 100.00% Hallucination Penalty Scores (HPS) across all feeding intensity states, confirming the complete eradication of fabricated environmental parameters and ungrounded statements. In evaluating catastrophic feeding error prevention, the critical Unsafe Continue Rate (USCR) demonstrated notable improvement. While the baseline LoRA configuration produced an unconstrained continuation judgment under safety critical withholding constraints yielding a USCR of 3.33% (1 out of 30 samples), counterfactual mDPO suppressed hazardous continuation commands, achieving a USCR of 0.00% (0 out of 30 samples) across all thirty evaluation scenarios. Detailed categorical distributions and safety diagnostics validating rule compliance are documented in Fig. C.2 (**Appendix C**). These empirical outcomes demonstrate that preference alignment enables the model to interpret subtle behavioral anomalies reliably and issue immediate feeding suspension commands to safeguard water quality.

### *3.5. Semantic fidelity, generation diversity, and decision interpretability*

Discrete control directives and explanatory rationales were evaluated independently. Discrete operational actions were assessed via categorical accuracy, while natural language metrics evaluated explanatory rationales against expert reference texts under greedy decoding (temperature = 0), as shown in Fig. 10. In Fig. 10(a), the standard Supervised Fine-Tuning baseline exhibited an asymmetric profile with high ROUGE-1 Recall (97.40%) alongside low Precision (22.70%), yielding ROUGE-1 and ROUGE-L F1 scores of 31.70% and 31.60% respectively, and a METEOR score of 58.10%. Counterfactual mDPO alignment improved lexical precision across generated rationales, raising ROUGE-1 Recall to 99.40% and ROUGE-1 Precision to 55.80%, which increased the ROUGE-L F1 score to 67.90% and the METEOR score to 85.30%. This gain confirms that preference alignment curtails redundant phrasing while preserving relevant diagnostic terminology. Precision-oriented BLEU evaluations in Fig. 10(b) showed consistent improvements. Counterfactual mDPO increased scores across n-gram orders, advancing BLEU-1 from 27.50% to 55.90% and BLEU-4 from 24.80% to 53.30%. The small score drop of 2.60% from 1-gram to 4-gram indicates higher n-gram precision across multi-word diagnostic phrases in generated rationales.

To measure generation redundancy, generation diversity on explanatory text was benchmarked in Table 3. Preference alignment improved lexical variety, elevating Distinct-3 from 6.68% to 7.81% while reducing output repetition, as Self-BLEU-2 and Self-BLEU-4 decreased to 52.88% and 49.07%, respectively. Class-wise Distinct-1 metrics showed consistent gains across all feeding categories, reaching 6.90% for weak, 8.99% for medium, and 5.46% for strong feeding regimes. Lower Self-BLEU values reflect reduced phrasing redundancy across distinct test cases rather than unconstrained action variation. Discrete operational directives remain strictly standardized.

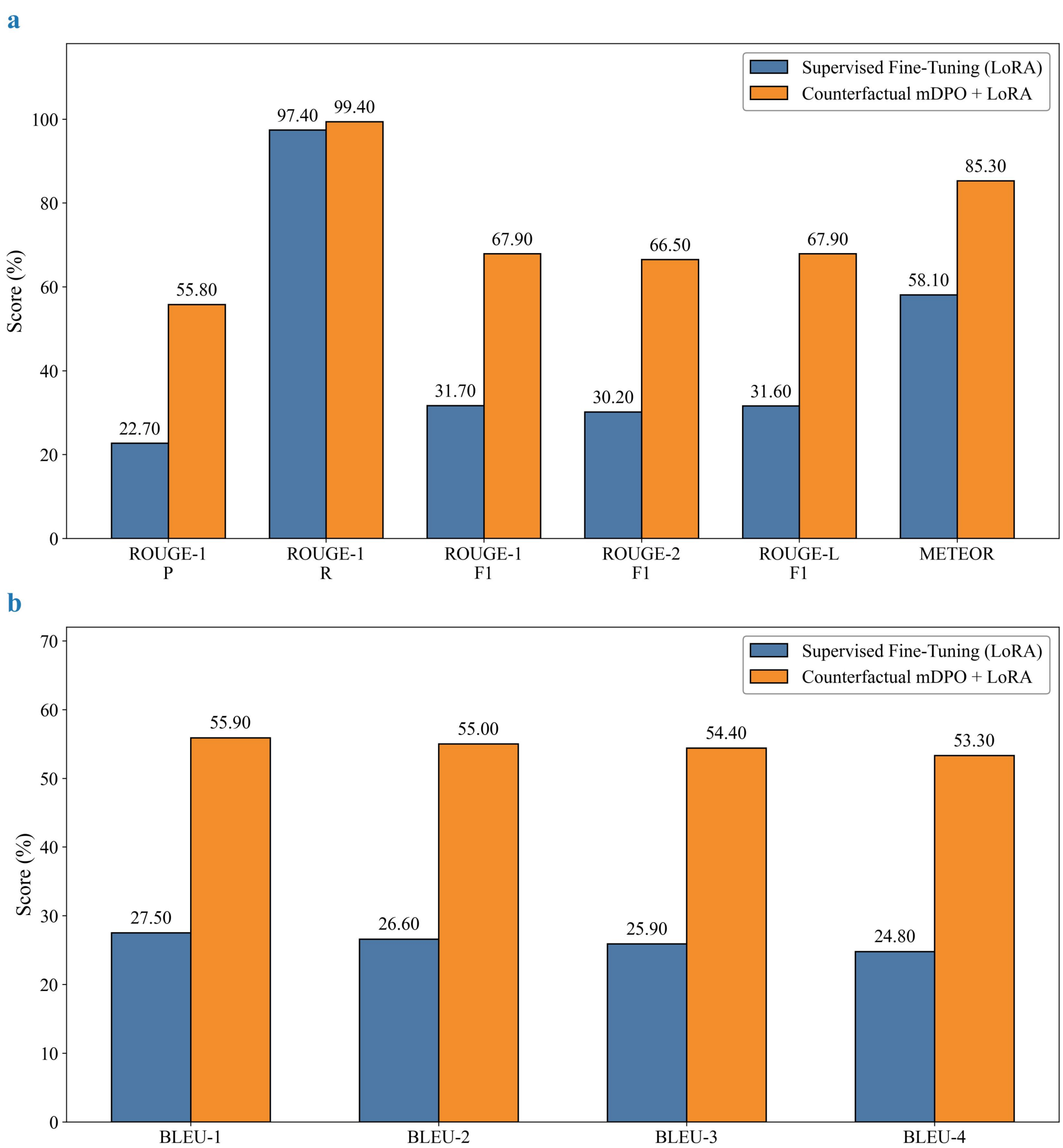

**Fig. 10.** Comparative evaluation of text generation quality on explanatory rationales between Supervised Fine-Tuning (LoRA) and Counterfactual mDPO. (a) Group-wise comparison of ROUGE precision, recall, F1, and METEOR scores. (b) Precision decay profiles across cumulative n-gram orders from BLEU-1 through BLEU-4.

**Table 3**

Quantitative comparison of lexical diversity and textual redundancy on explanatory rationales between LoRA and counterfactual mDPO across lexical, structural, and class-wise metrics.

| Optimization Strategy | Distinct-1 (%) | Distinct-2 (%) | Distinct-3 (%) | Self-BLEU-2 (%) | Self-BLEU-4 (%) | Distinct-1 (%) | | |
|---|---|---|---|---|---|---|---|---|
| | | | | | | Weak | Medium | Strong |
| Supervised Fine-Tuning (LoRA) | 2.96 | 5.57 | 6.68 | 58.79 | 52.92 | 5.84 | 8.27 | 5.43 |
| Counterfactual mDPO + LoRA | 3.47 | 6.57 | 7.81 | 52.88 | 49.07 | 6.90 | 8.99 | 5.46 |

A qualitative comparative evaluation substantiated decision fidelity between Supervised Fine-Tuning (LoRA) and counterfactual mDPO under identical behavioral and contextual inputs. As illustrated in the inference transcripts in Fig. C.1, when presented with a strong feeding scenario characterized by an AC of 0.76, swimming speed of 19.68 cm/s, and acceleration of 543.97 cm/s$^2$, the baseline Supervised Fine-Tuning policy exhibits an operational discrepancy. Specifically, the baseline model generates text centered on post-suspension water quality monitoring and conservatively predicts an extended 7-second feeding interval rather than the required 5-second interval. This interval pacing mismatch reflects an operational error under vigorous feeding conditions. In contrast, the counterfactual mDPO policy associates high-magnitude visual kinematics with active feeding guidance, prescribing the 5-second feeding interval alongside continuous feeder operation and full ration delivery. This outcome confirms that counterfactual preference alignment reduces conservative pacing errors and improves alignment between continuous visual kinematics and discrete actuator execution (Fig. C.1, **Appendix C**).

## 4. Discussion

### *4.1. Physical representation of collective feeding activity via AC*

The AC integrates disparate kinematic variables into a traceable and comparable metric of collective motion. In this study, the mean-based AC demonstrated reliable capability in distinguishing feeding intensity regimes. Constructed from explicit kinematic variables, the metric attributes overall activity shifts to specific motion components. This provides a physically grounded basis for partitioning feeding regimes and guiding downstream behavioral modeling. Furthermore, using the AC as a macroscopic baseline reveals kinematic heterogeneity across the school. Under comparable AC values, fish schools can exhibit either a heterogeneous pattern where a few individuals are highly active while others remain quiescent, or a homogeneous pattern characterized by uniform swimming speeds across the population. Because mean-based macroscopic metrics are insensitive to internal velocity distributions, aggregate quantification and fine-grained behavioral encoding exhibit an essential complementary relationship. Prior research demonstrated the utility of parsing behavioral structure from continuous kinematic data by linking keypoint tracking to pose dynamics (Weinreb et al., 2024). In alignment with this methodology, the subsequent modules in THPL retain the AC as a macroscopic activity baseline while incorporating temporal trajectory dynamics and spatial distributions into the HBE. This architecture supplies richer behavioral evidence for final decision-making than a single scalar index.

### *4.2. Joint modeling of kinematic feeding heterogeneity and collective relationships via HBE*

In dense fish schools, severe occlusions alter the count of trackable targets. Furthermore, premature spatial averaging can obscure kinematic variance among individuals. Multi-animal tracking studies have identified occlusions, proximal interactions, and individual associations as central challenges in behavioral quantification (Lauer et al., 2022). The HBE addresses these issues through a decoupled structure that processes individual temporal dynamics before collective set aggregation. This design allows transient motion patterns to be captured prior to population-level pooling. The framework preserves individual temporal continuity alongside collective dynamic changes. It also retains transient feeding cues such as burst swimming and rapid turning. In addition, this design reduces dependence on persistent individual identities, thereby improving model applicability under intermittent tracking and identity switches.

The dual-evidence representation counteracts the over-compression of collective dynamics inherent in average metrics. It enables the model to differentiate behavioral states that exhibit similar macro-level intensity, such as localized feeding competition versus generalized schooling activity. The superior performance of combining explicit and implicit tokens confirms that both representations provide complementary perspectives on school dynamics. The former grounds motion features in traceable physical properties. The latter supplements spatiotemporal variations that summary statistics cannot capture. Feeding both representations into the decision layer preserves macroscopic physical meaning while avoiding the erasure of individual variation. These results demonstrate that the HBE provides a representational foundation for analyzing scenarios with comparable aggregate activity but distinct internal velocity distributions. Future work can further separate states such as localized scramble feeding and uniform swimming using targeted ground-truth annotations.

### *4.3. Contextual decision integration via LLM and mDPO preference alignment*

In THPL, the LLM functions as a semantic decision interface. It synthesizes behavioral evidence, environmental telemetry, husbandry background, and expert rules into structured feeding directives. Practical feeding management requires more than classifying feeding intensity. It demands coordinated determination of feeder states, feeding intervals, feed rations, and operational rationales. The fine-tuned LLM performs high-level synthesis and conversion rather than direct processing of raw visual pixels. For standard classification of feeding intensity alone, conventional classifiers or rule-based models remain more direct and computationally efficient. However, when decisions require unifying multi-source inputs and generating actionable, interpretable guidance for farm operators, the context integration and instruction-following abilities of the LLM provide distinct utility. Prior studies have highlighted the capacity of LLMs to bridge heterogeneous inputs with operational execution. Boiko et al. (2023) linked LLMs with literature retrieval, code execution, and hardware automation for experimental planning. Mon-Williams et al. (2025) integrated knowledge retrieval, vision, and haptic feedback for robotic manipulation. In contrast to these chemical and robotic domains, this study focuses on integrating biological state representations with aquaculture operational guidelines during feeding progression.

Counterfactual mDPO supports the alignment of behavioral evidence with operational constraints. This method translates expert domain knowledge into pairwise response preferences, embedding rules into model parameters rather than relying on prompts alone. This mechanism should be understood as learning contextual policy selection rather than uncovering causal biological laws. The improved decision accuracy after mDPO demonstrates that preference optimization reshapes decision boundaries and error distributions. Its practical utility depends on the real-world operational costs of different error types. Regarding text metrics, aligned outputs show higher lexical overlap with expert references and reduced repetition, facilitating structured, verifiable explanations. THPL incorporates text quality, evidence grounding, and action compliance into a joint evaluation. Combining preference alignment with an independent output validator maintains conversational flexibility while enforcing deterministic operational boundaries (Steyvers et al., 2025).

### *4.4. Limitations and future research directions*

Several limitations in this study warrant consideration. First, all video recordings were acquired from a single cohort of *O. mykiss*. The temporal test split evaluates only within-cohort temporal generalization and cannot substitute for independent biological replicates. Second, AC boundaries and kinematic features can vary with fish body size, stocking density, and individual traits. Their stability and calibration requirements across independent cohorts require systematic verification. Third, the pre-configured perturbation tests do not capture continuous online performance under concurrent tracking errors and dynamic environmental fluctuations. Finally, the sample size of the test set remains modest. Current decision metrics evaluate policy compliance rather than demonstrating empirical improvements in feed conversion ratio, growth rate, water quality, production cost, or animal welfare. Future investigations should incorporate multi-cohort trials and independent biological replicates, validating real-world production benefits through extended closed-loop feeding trials. Additionally, integrating uncertainty estimation, out-of-distribution detection, human-in-the-loop overrides, and fail-safe controls will be critical to evaluating operational stability and safety boundaries in continuous deployment. Further studies could leverage unseen environmental conditions and fine-grained behavioral annotations to validate the utility of contextual decision-making and motion heterogeneity.

## 5. Conclusion

This study investigated a behavior-grounded generative framework designed to translate fish motion into interpretable feeding decisions within recirculating aquaculture systems. The framework integrates a dimensionless AC, a HBE, and an LLM policy optimized via counterfactual preference alignment. The results indicate that THPL consolidates multidimensional kinematics into physically grounded behavioral priors while extracting structured behavioral evidence from fragmented trajectories. On this foundation, preference alignment enables the model to generate context-adaptive feeding directives that comply with expert operational rules. This provides a methodology for mapping continuous visual kinematics to verifiable control decisions. Despite these findings, this study is subject to limitations, including reliance on a single fish cohort, limited biological replication, and the absence of closed-loop operational validation across diverse farming conditions. Future work should focus on cross-cohort calibration, uncertainty-aware safety mechanisms, and long-term production trials to advance behavior-grounded generative decision support toward practical implementation in commercial aquaculture.

**Acknowledgments**

This research was supported by National key R&D Program of China (2024YFD2400100), the scientific research project in Fengjiawan of Hainan Provincial Seed Industry Laboratory (B24H10035), the Fundamental Research Funds for the Central Universities (226-2025-00074) and China Agriculture Research System of MOF and MARA (CARS-49).

**CRediT authorship contribution statement**

Meng Liang: Conceptualization, Methodology, Formal analysis, Software, Visualization, Investigation, Writing - original draft, Writing - review & editing. Guanbo Feng: Methodology, Formal analysis, Visualization, Writing - review & editing. Haozhuang Chi: Methodology, Writing - review & editing. Shilong Zhao: Resources, Data curation. Zhixin Xiong: Investigation. Yuhang He: Writing - review & editing. Wenfeng Han: Visualization, Writing - review & editing. Tianhao Zhao: Writing - review & editing. Zhihong Ma: Supervision, Project administration, Conceptualization, Resources, Writing - review & editing. Ying Liu: Conceptualization, Project administration, Funding acquisition, Supervision, Writing - review & editing.

**Declaration of competing interest**

The authors declare that they have no known competing financial interests or personal relationships that could have appeared to influence the work reported in this paper.

**Data and code availability**

The data and code in this study are available from the author upon reasonable academic request.

## Appendix A. Validation of kinematic data pipeline and behavioral indices

### A.1. Physical-consistent kinematic augmentation and statistical distributional fidelity verification

A physical-consistent kinematic augmentation (PCKA) pipeline was applied to the 239 raw training segments, generating 2,868 transformed kinematic views to construct a composite training corpus of 3,107 instances (Section 2.1.2 and Table A.1). This pipeline incorporated spatial rigid flipping, global spatial rotation, viewpoint affine shearing, and temporal resampling to eliminate spatial biases. Spatial coordinates underwent geometric transformations, while planar velocities and accelerations were strictly re-calculated from transformed coordinates using finite differences to maintain internal kinematic consistency. To evaluate whether transformed views retain empirical kinematic distributions, multi-variable correlation distributions and two-sample Kolmogorov-Smirnov tests were conducted for displacement, velocity, acceleration, and the Activity Coefficient (AC) across all feeding intensity classes.

**Table A.1**

Summary of spatio-temporal data augmentation strategies, implementation details, and sample distributions.

| Augmentation Method | Transformation Details | Target Objective | Sample Count (N) |
|---|---|---|---|
| Raw Baseline | Original captured spatio-temporal sequences | Unaugmented experimental baseline | 239 |
| Spatial Rigid Flipping | Horizontal, vertical, and diagonal mirror transformations | Eliminate spatial positional bias | 717 (239×3) |
| Global Spatial Rotation | Isotropic coordinate rotations (90º, 180º, 270º) | Remove directional preferences | 717 (239×3) |
| Viewpoint Affine Shearing | Global shearing transformations along spatial axes | Simulate camera tilt and water refraction | 717 (239×3) |
| Temporal Resampling | Cubic spline interpolation along the time axis | Simulate rhythm variations ($C^2$ continuity) | 717 (239×3) |
| Total Dataset | Comprehensive multi-view spatio-temporal expansion | Balanced dataset for robust model training | 3,107 |

Note: Each augmentation pipeline produces 717 synthesized samples, resulting in 2,868 augmented instances and a final dataset size of 3,107 sequences.

The bivariate scatter matrices and diagonal kernel density estimation curves for weak feeding in Fig. A.1(a), medium feeding in Fig. A.1(b), and strong feeding in Fig. A.1(c) illustrate high statistical concordance between the raw trajectories and synthesized views. Across all three feeding intensity categories, the diagonal kernel density distributions of displacement, velocity, acceleration, and the AC exhibit substantial overlap between raw and augmented data. Furthermore, the off-diagonal bivariate scatter distributions confirm that empirical linear and non-linear correlations among swimming displacement, velocity, and burst acceleration are largely preserved across the kinematic space. This demonstrates that the augmentation operations enriched training diversity without introducing severe numerical distortions or extreme outliers into the dataset.

The quantitative statistical similarity was substantiated by the two-sample Kolmogorov-Smirnov tests presented in Fig. A.2. All calculated p-values across displacement, velocity, acceleration, and the composite AC exceed the threshold of 0.05 across all three feeding intensity classes, ranging from 0.23 to 1.00. These non-parametric test results indicate that transformed sequences introduced no severe statistical distribution drift in low-order kinematics. This confirms empirical distributional consistency across training views, providing a stable data foundation for training the HBE.

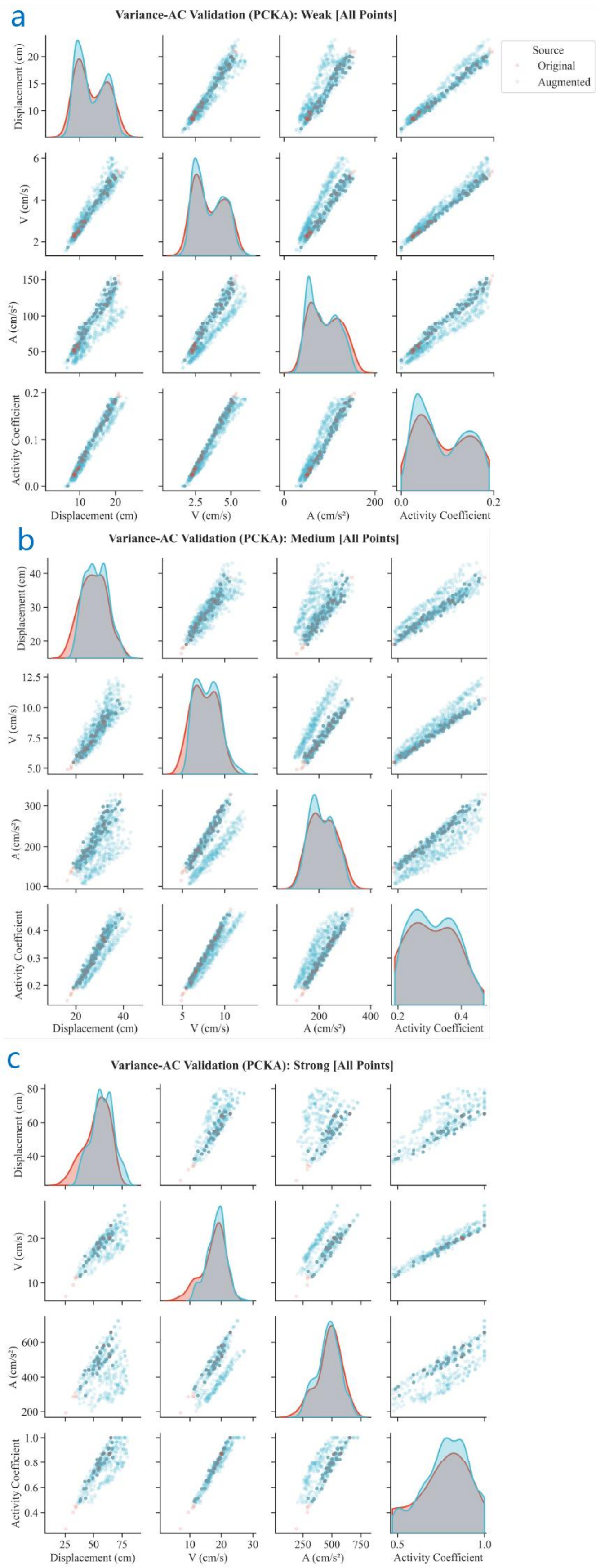

**Fig. A.1.** Kinematic correlation scatter matrices and density distributions of original versus PCKA augmented data across feeding intensity classes: (a) weak feeding, (b) medium feeding, and (c) strong feeding.

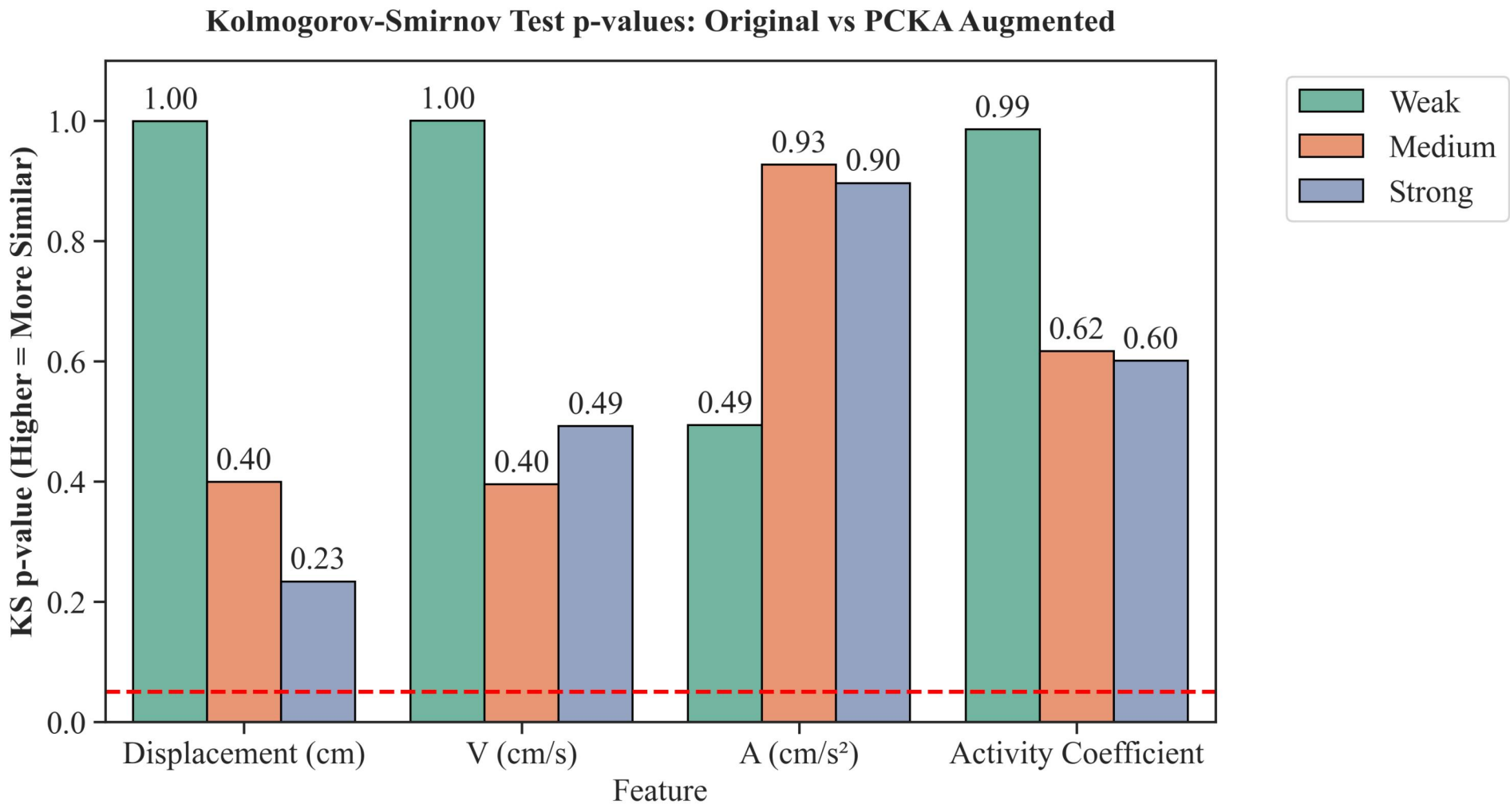

**Fig. A.2.** Two-sample Kolmogorov-Smirnov test p-values between original and PCKA augmented distributions across kinematic features and feeding intensity classes.

### A.2. Formulation ablation and discriminatory evaluation of candidate Activity Coefficients

An ablation study comparing the standardized mean ( $AC_{mean}$ ) formulated in Eq. (A.1), variance ( $AC_{var}$ ) expressed in Eq. (A.2), and magnitude energy ( $AC_{energy}$ ) defined in Eq. (A.3) was conducted to determine the optimal mathematical representation for physical behavior aggregation.

$$AC_{mean} = \frac{1}{3}\left(\overline{Z}^{D} + \overline{Z}^{V} + \overline{Z}^{A}\right) \tag{A.1}$$

$$AC_{var} = \frac{1}{3}\left(\mathrm{Var}(Z^{D}) + \mathrm{Var}(Z^{V}) + \mathrm{Var}(Z^{A})\right) \tag{A.2}$$

$$AC_{energy} = \sqrt{\frac{1}{3}\left((\overline{Z}^{D})^2 + (\overline{Z}^{V})^2 + (\overline{Z}^{A})^2\right)} \tag{A.3}$$

Where $\overline{Z}^{D}$, $\overline{Z}^{V}$, and $\overline{Z}^{A}$ represent the school-averaged standardized displacement, velocity, and acceleration, and $\mathrm{Var}(\cdot)$ denotes the intra-group variance across tracked individuals within the segment.

A comparative ablation among $AC_{mean}$, $AC_{var}$, and $AC_{energy}$ evaluated candidate formulations for the macroscopic behavior index on the independent validation split ($N$=30, comprising 10 balanced sequences per intensity class) prior to model freezing (Figs. A.3 to A.6). As shown in the box plots (Fig. A.3) and summary metrics (Fig. A.4), $AC_{mean}$ achieved the highest between-group discriminatory capability. A one-way ANOVA yielded an F-statistic of $F$ (2, 27) = 176.88 ($p$ < 0.001), an effect size $\eta^2$ of 0.9291, and an empirical classification accuracy of 1.0000 across weak (mean = -0.4098), medium (mean = -0.0797), and strong (mean = 0.4895) feeding classes. In contrast, $AC_{var}$ exhibited moderate discrimination (accuracy 0.9000, $F$ (2, 27) = 59.07, $\eta^2$ = 0.8140). $AC_{energy}$ suffered severe performance degradation (accuracy 0.6333, $F\,(2, 27) = 41.03$, $\eta^2 = 0.7524$). As visualized in the density distributions and pairwise scatter matrices in Fig. A.5, $AC_{energy}$ degraded separation because squaring zero-centered standardized variables folded negative velocity deviations into positive quadratic magnitudes. This sign folding forced weak and strong feeding states to overlap in the energy space, reflecting a structural limitation of standardized quadratic amplitudes rather than an intrinsic failure of energy metrics. Finally, the pairwise effect size heatmap in Fig. A.6 confirms that $AC_{mean}$ produced substantial absolute Cohen's d values (3.548 for weak vs. medium, 8.221 for weak vs. strong, and 4.731 for medium vs. strong). Consequently, $AC_{mean}$ was selected as the optimal kinematic aggregation metric.

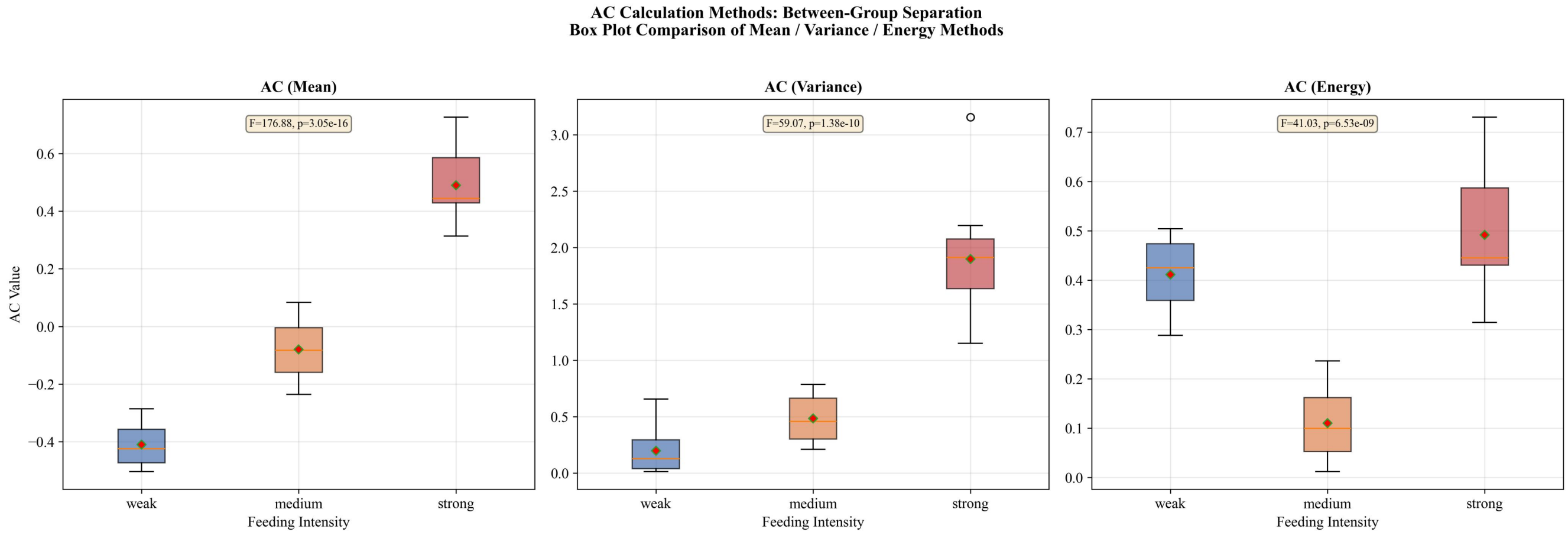

**Fig. A.3.** Box plot distributions and between-group separation metrics for candidate Activity Coefficient formulations across feeding intensity classes.

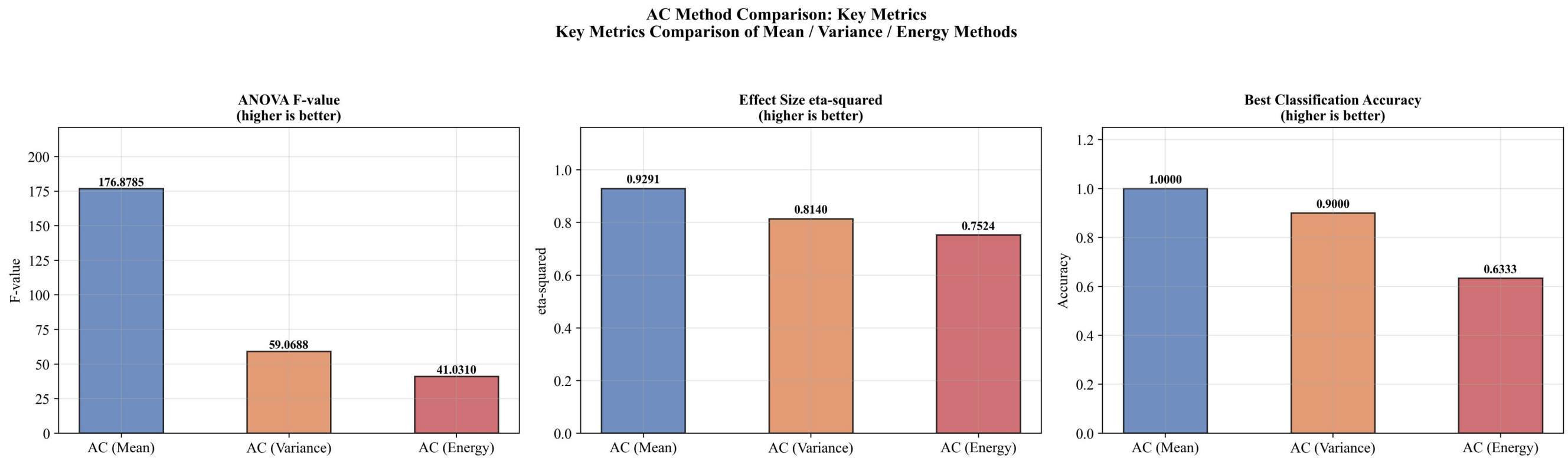

**Fig. A.4.** Key metrics comparison including ANOVA F-value, effect size eta-squared, and optimal classification accuracy.

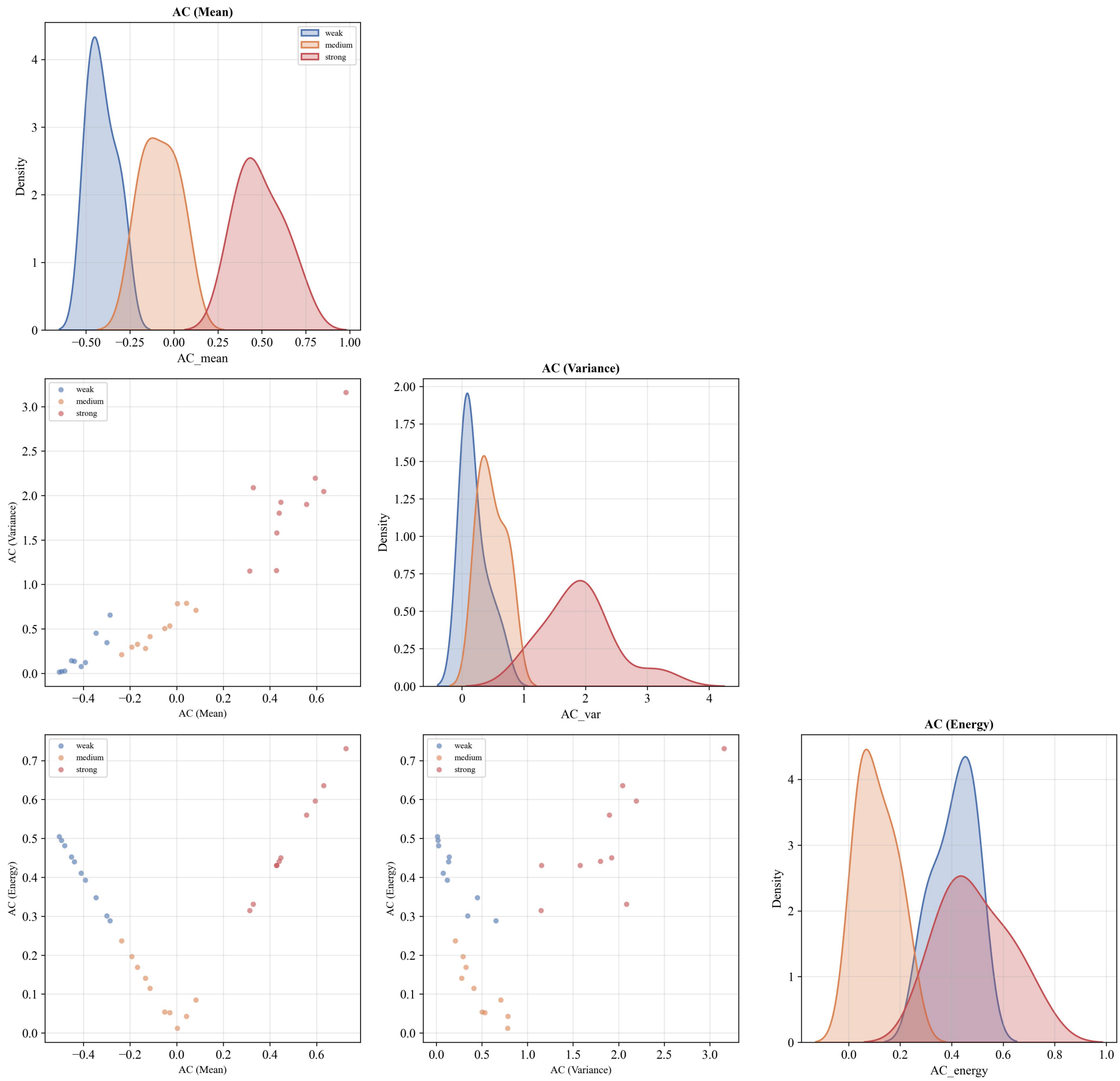

**Fig. A.5.** Pairwise scatter matrices and kernel density distributions comparing the three AC formulations.

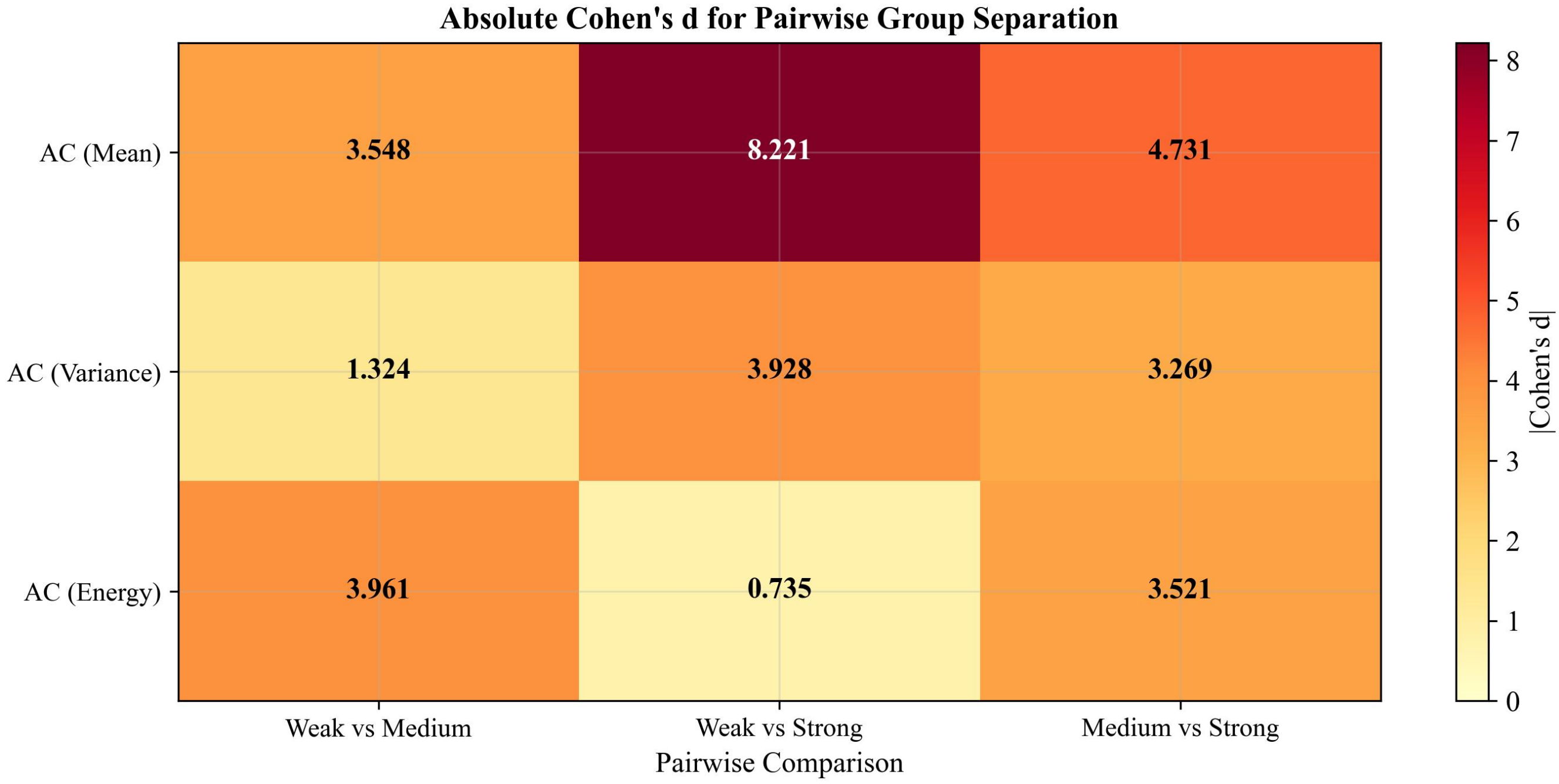

**Fig. A.6.** Pairwise Cohen's d effect size heatmap and bivariate correlation scatter matrices comparing AC(Mean), AC(Variance), and AC(Energy) formulations.

### A.3. Data-driven empirical threshold optimization and decision boundary determination

An optimal empirical thresholding method was established to map continuous AC values into discrete feeding intensity categories. Let $x$ denote the scalar AC value, and let $\theta_W$ and $\theta_M$ represent the decision boundaries for weak and medium feeding regimes, respectively. The optimal threshold pair ($\theta_W^*$, $\theta_M^*$) is determined by maximizing the empirical classification accuracy over the dataset of $N$ samples, subject to the monotonic physical constraint $\theta_W < \theta_M$. A grid search optimization with a step size of $\Delta$ = 0.001 was executed across candidate boundaries to identify coordinates yielding the highest classification accuracy. Formally, the piecewise classification function, accuracy objective, and optimal threshold formulation are expressed in Eqs. (A.4)-(A.6).

$$f(x) = \begin{cases} \text{Weak}, & 0 \le x \le \theta_W \\ \text{Medium}, & \theta_W < x \le \theta_M \\ \text{Strong}, & \theta_M < x \le 1 \end{cases} \tag{A.4}$$

$$\text{Accuracy}(\theta_W, \theta_M) = \frac{1}{N}\sum_{i=1}^{N} \mathbf{1}[f(x_i) = y_i] \tag{A.5}$$

$$(\theta_W^*, \theta_M^*) = \arg\max_{\theta_W, \theta_M} \text{Accuracy}(\theta_W, \theta_M) \tag{A.6}$$

**A.4. Ecological and biomechanical validation of the negative correlation between local fish density and feeding activity**

Spatial schooling distribution was quantified via mean Local Density, defined as the inverse of the mean nearest-neighbor distance across tracked individuals (contrasting with the absolute neighborhood counts in Fig. 7). Across the 299 raw benchmark sequences, Local density showed a significant negative monotonic correlation with the composite AC (Spearman $r$ = -0.9629, $p < 0.001$, Pearson $r$ = -0.8783, $p < 0.001$). Because both metrics derive from the trajectory tensor $\mathbf{X}$, this correlation reflects behavioral consistency between spatial dispersion and locomotion activity during feeding, rather than an independent disproof of tracking artifacts. As visualized in the bivariate heatmap in Fig. A.7(a), sample concentrations shift continuously from high-density, low-AC resting zones toward low-density, high-AC foraging regimes across the decision boundaries (0.181 and 0.462). Furthermore, a Chi-square test of independence across five density intervals confirmed a significant structural shift in schooling organization ($\chi^2 = 426.82$, $p < 0.001$, Fig. A.7(b)), capturing the biological transition from cohesive shoaling at rest to spatial dispersal during competitive food scrambles.

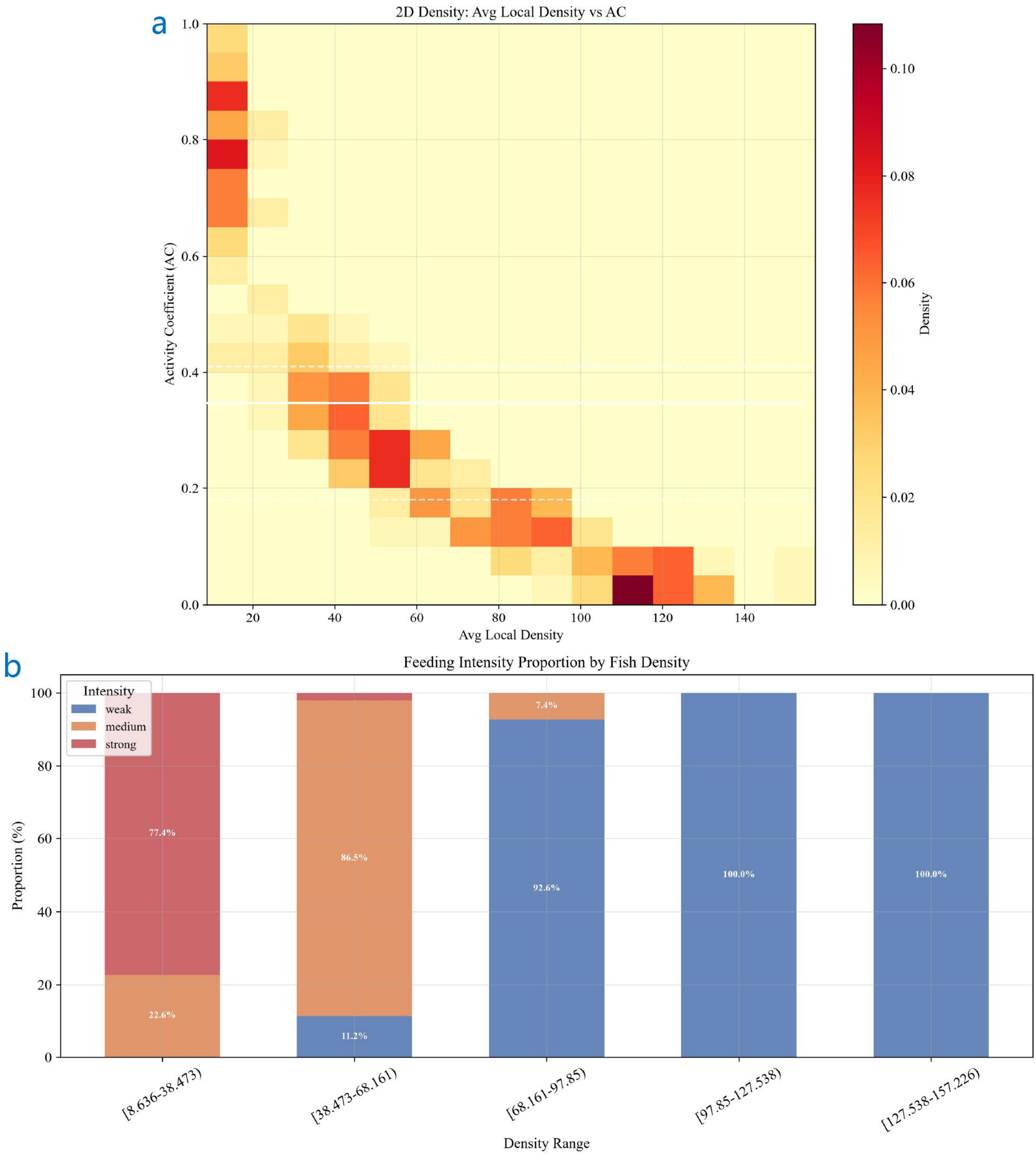

**Fig. A.7.** Quantitative correlation between mean local fish school density and feeding intensity across the 299 raw benchmark sequences. Subplot (a) shows the bivariate density heatmap of normalized local density versus AC with supervisory boundaries of 0.181 and 0.462; subplot (b) shows feeding intensity class proportions across five progressive local density intervals.

## A.5. Representation probing of kinematic velocity heterogeneity

While the scalar AC effectively quantifies collective swimming vigor, mean based spatial aggregation inherently compresses the underlying velocity distribution. Consequently, identical AC values may correspond to either a non-uniform distribution where a subset of individuals exhibits high velocities alongside slower moving peers, or a uniform distribution where the majority of individuals swim at homogeneous intermediate velocities. To verify whether downstream behavioral tokens preserve this kinematic heterogeneity rather than degenerating into macroscopic activity proxies, linear probing evaluations were conducted across the thirty held out benchmark sequences.

Kinematic dispersion was parameterized by the coefficient of variation (CV) of per fish mean swimming speed ( $CV = \sigma_v / \bar{v}$ ), derived from individual tracking trajectories across the ten temporal segments. To probe behavioral representations without introducing unconstrained parameters, frozen token vectors were evaluated using ridge regression probes. For each sequence, the 14 behavioral tokens ($M$=4 explicit and $K$=10 implicit soft tokens) were mean pooled across the token dimension into a 4096 dimensional feature vector, avoiding excessive dimensionality expansion. Evaluations followed a strict five-fold cross validation protocol (random seed 42). Within each cross validation split, feature standardization via z score scaling and Ridge regressor fitting ($\alpha$ = 1.0) were fitted exclusively on the training folds and subsequently evaluated on the out of fold test split to ensure zero data leakage. The HBE backbone remained entirely frozen and had received no supervision regarding velocity dispersion during prior training stages.

Across five-fold evaluation, linear probing on full behavioral tokens achieved a cross validated determination coefficient of $R^2$ = 0.851 (Pearson $r$ = 0.923, $p$ < 0.001) for predicting speed CV, while simultaneously capturing overall activity ($R^2$ = 0.894 for AC, Pearson $r$ = 0.951, $p$ < 0.001). These empirical results demonstrate that the learned token space linearly preserves statistical dispersion in individual swimming speeds beyond population averages. However, this evaluation strictly confirms representation level retention of kinematic variance under the specified probing protocol. These findings do not imply verified behavioral hierarchy, feed intake dominance, or competitive social structures, which require fine scale nutritional validation and individual feeding ground truth in future research.

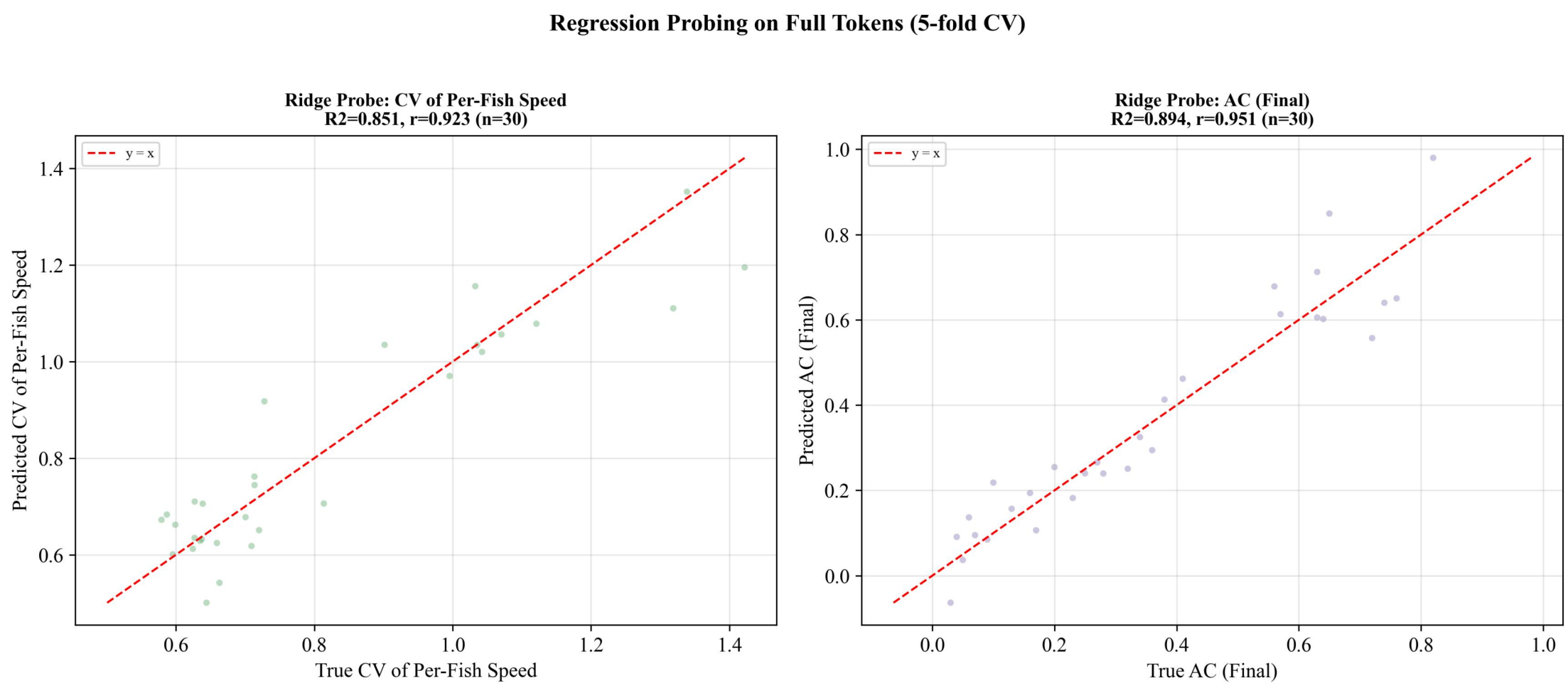

**Fig. A.8.** Linear regression probing on full behavioral tokens across thirty experimental sequences. The left panel plots predicted against empirical coefficient of variation values for per-fish swimming speed, demonstrating distribution shape preservation. Correspondingly, the right panel depicts model fidelity

against the macroscopic AC to substantiate global activity level retention.

a

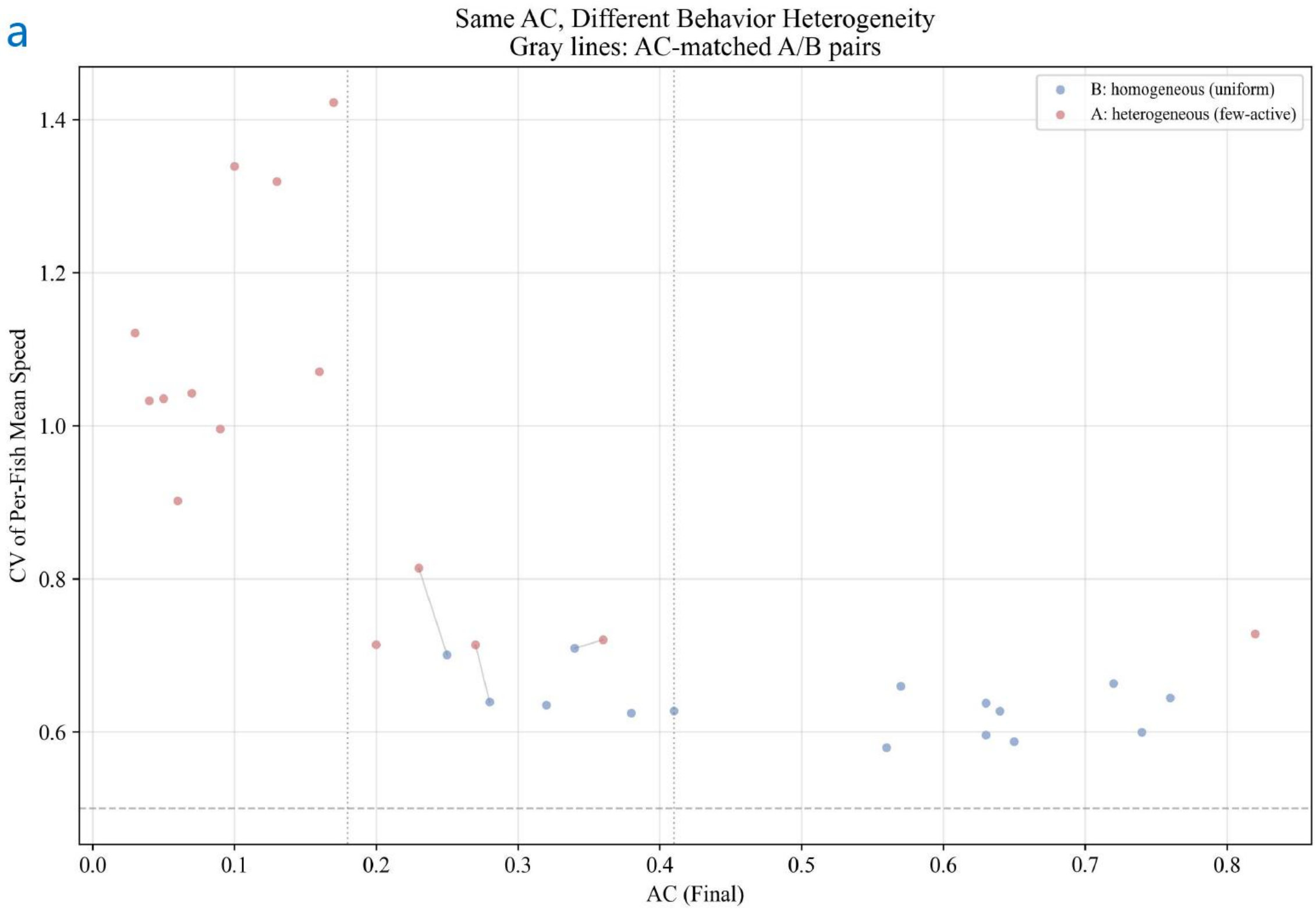

b

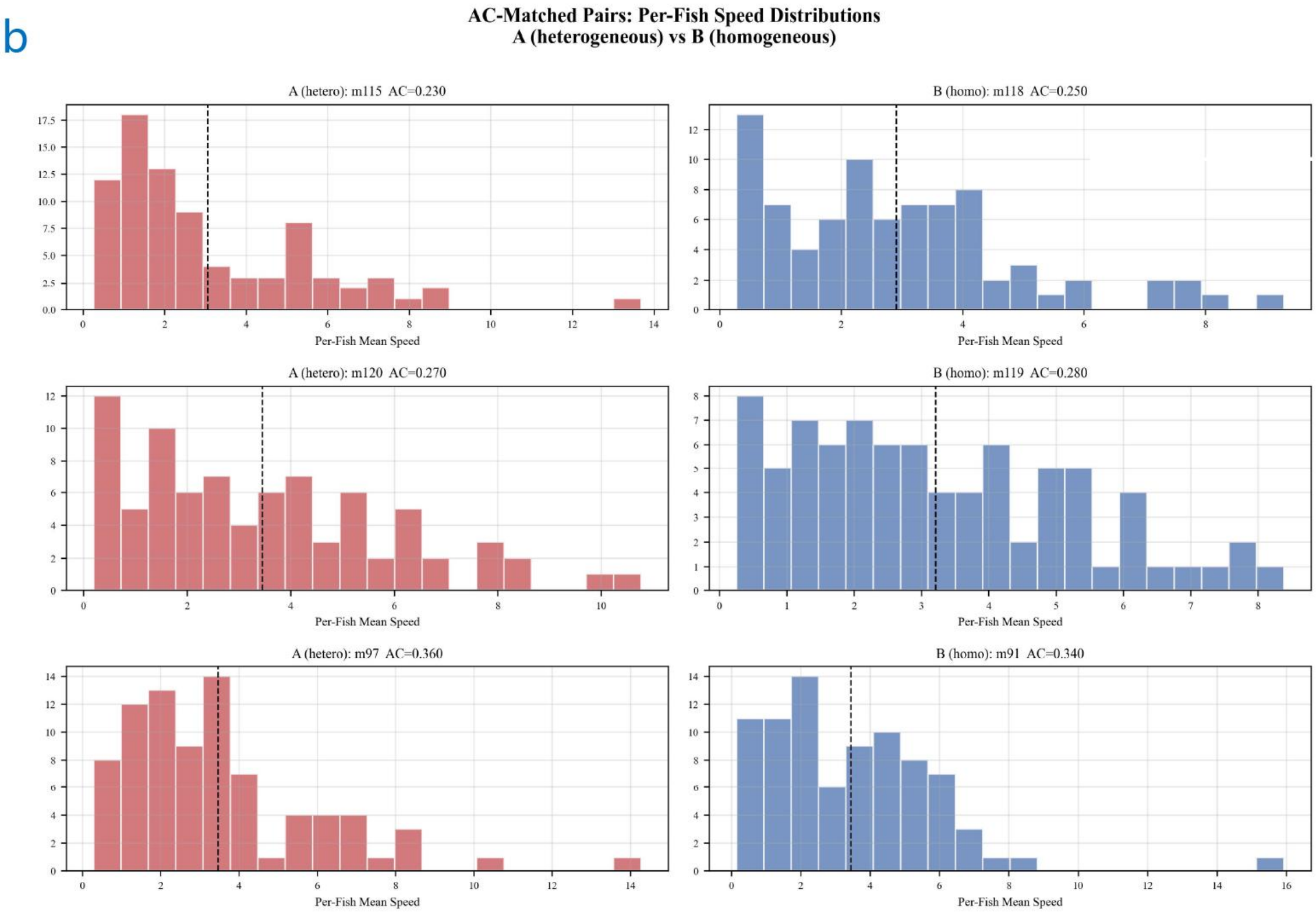

**Fig. A.9.** Analysis of fish schooling behavioral heterogeneity across identical AC regimes. Subplot (a) shows the scatter distribution of AC versus the Coefficient of Variation of per-fish mean speed, with connected pairs denoting matched macroscopic activity levels across heterogeneous and homogeneous patterns. Subplot (b) displays per-fish speed distribution comparisons across three representative matched pairs, contrasting skewed minority-active distributions with uniform moderate-swimming distributions under equivalent AC baselines.

**Appendix B. Mathematical formulations of architectural blocks, losses, and evaluation metrics**

**B.1. Architectural operator formulations of HBE**

The fundamental mathematical operator driving both individual temporal modeling and collective set-level aggregation in the Hierarchical Behavior Encoder (HBE) is the scaled dot-product attention mechanism (Vaswani et al., 2017), which dynamically scales values **V** based on queries **Q** and keys **K** of dimension $d_{attn}$ as defined in Eq. (B.1). To model multi-fish interactions across entity subspaces without imposing an artificial spatial order, the Set Transformer architecture defines the Multihead Attention Block (MAB) for two representation sets **X** and **Y** (Lee et al., 2019). This structure computes an intermediate normalized attention feature **H** via Eq. (B.2) and passes it through a row-wise feedforward network (rFF) with residual connection to yield the final block output formulated in Eq. (B.3). Building upon this formulation, the Set Attention Block (SAB) is defined as self-attention over the same set via Eq. (B.4). Within the HBE pipeline, the temporal modeling layer applies multi-head self-attention driven by Eq. (B.1) across the temporal dimension $T$ for each individual fish sequence to produce entity representations $\mathbf{H}_{temp}$, while the subsequent collective layer applies SAB across the entity dimension $N$ to yield the permutation-invariant school representation $\mathbf{H}_{set}$.

$$\text{Attention}(\mathbf{Q},\mathbf{K},\mathbf{V}) = \text{Softmax}\left(\frac{\mathbf{QK}^T}{\sqrt{d_{attn}}}\right)\mathbf{V} \tag{B.1}$$

$$\mathbf{H} = \text{LayerNorm}(\mathbf{X} + \text{Multihead}(\mathbf{X},\mathbf{Y},\mathbf{Y})) \tag{B.2}$$

$$\text{MAB}(\mathbf{X},\mathbf{Y}) = \text{LayerNorm}(\mathbf{H} + \text{rFF}(\mathbf{H})) \tag{B.3}$$

$$\text{SAB}(\mathbf{X}) = \text{MAB}(\mathbf{X},\mathbf{X}) \tag{B.4}$$

**B.2. Parameter-efficient fine-tuning formulation via LoRA**

Low-Rank Adaptation (LoRA) is integrated into the multi-head attention projection layers to adapt the Large Language Model (LLM) to aquaculture feeding management without incurring prohibitive computational overhead (Hu et al., 2021). The pre-trained parameter matrix $\mathbf{W}_0 \in \mathbb{R}^{d\times k}$ is frozen during training, and the intrinsic rank constraint is modeled by injecting trainable low-rank decomposition matrices $A$ and $B$. The modified forward pass computation is mathematically formulated in Eq. (B.5), where $x \in \mathbb{R}^k$ and $h \in \mathbb{R}^d$ represent the layer input and output vectors respectively, $\mathbf{B} \in \mathbb{R}^{d\times r}$ and $\mathbf{A} \in \mathbb{R}^{r\times k}$ denote the low-rank adapter matrices with rank $r \ll \min(d,k)$ serves as a constant scaling factor:

$$h = W_0x + \Delta Wx = W_0x + \frac{\alpha}{r}BAx \tag{B.5}$$

### B.3. Standard evaluation metric formulations

Accuracy is defined as the proportion of model-generated structured feeding decisions that are entirely consistent with the expert baseline strategies, computed via Eq. (B.6), where $total_{correct}$ represents the number of samples with all three decision fields (feeder state, feeding interval, and feed amount) predicted correctly, and $total$ is the total number of samples. To evaluate class-wise performance, the decision recall measures the coverage of correct predictions within specific intensity classes. As formulated in Eq. (B.7), $Recall_c$ is calculated where $correct_c$ represents the correctly predicted samples for class $c$ and $n_c$ is the actual total number of samples in class $c \in \{weak, medium, strong\}$. The Invalid Rate quantifies the proportion of model outputs that suffer from formatting collapse or lack analysable decision fields, expressed in Eq. (B.8), where $N_{invalid}$ represents the number of samples missing at least one decision field.

$$Accuracy = \frac{total_{correct}}{total} \times 100\% \tag{B.6}$$

$$Recall_c = \frac{correct_c}{n_c} \times 100\% \tag{B.7}$$

$$Invalid\ Rate = \frac{N_{invalid}}{N_{total}} \times 100\% \tag{B.8}$$

Semantic fidelity is quantified through BLEU, which evaluates the exact n-gram overlap between generated rationales and expert reference texts via Eq. (B.9). Extending this to measure lexical diversity, Self-BLEU evaluates pairwise n-gram overlap across the generated responses according to Eq. (B.10). In both formulations, BP denotes the brevity penalty, with n-gram precision denoted as $p_n$ for external reference evaluations and $p_n(x_i, X_{-i})$ for internal corpus evaluations. A lower Self-BLEU score indicates reduced phrasing redundancy across distinct test cases. ROUGE-L evaluates structural lexical overlap based on the Longest Common Subsequence (LCS), measuring surface sentence similarity between generated explanations and reference texts. The precision ( $P_{LCS}$ ) and recall ( $R_{LCS}$ ) of ROUGE-L are formulated in Eq. (B.11), with the F1 score calculated via Eq. (B.12). METEOR incorporates stemming and synonym matching to evaluate semantic alignment beyond strict surface matching, computing the harmonic mean ( $F_{mean}$ ) in Eq. (B.13) before applying a fragmentation penalty in Eq. (B.14). Furthermore, BERTScore computes cosine similarity over contextual token embeddings to evaluate semantic similarity in latent space. Precision and recall for BERTScore are defined in Eq. (B.15), and the composite F1 score is formulated in Eq. (B.16).

$$BLEU = BP \cdot \exp\left( \sum_{n=1}^{N} w_n \log p_n \right) \tag{B.9}$$

$$Self - BLEU = \frac{1}{M} \sum_{i=1}^{M} BP \cdot \exp\left( \sum_{n=1}^{N} w_n \log p_n(x_i, X_{-i}) \right) \tag{B.10}$$

$$R_{LCS} = \frac{LCS(X,Y)}{m}, \quad P_{LCS} = \frac{LCS(X,Y)}{n} \tag{B.11}$$

$$F_{LCS} = \frac{(1+\beta^2) R_{LCS} P_{LCS}}{R_{LCS} + \beta^2 P_{LCS}} \tag{B.12}$$

$$F_{mean} = \frac{10PR}{R + 9P} \tag{B.13}$$

$$METEOR = F_{mean} \cdot \left( 1 - 0.5 \left( \frac{chunks}{unigrams_{matched}} \right)^3 \right) \tag{B.14}$$

$$R_{BERT} = \frac{1}{|x|} \sum_{x_i \in x} \max_{y_j \in y} \mathbf{x}_i^T \mathbf{y}_j, \quad P_{BERT} = \frac{1}{|y|} \sum_{y_j \in y} \max_{x_i \in x} \mathbf{x}_i^T \mathbf{y}_j \tag{B.15}$$

$$F_{BERT} = 2 \frac{P_{BERT} \cdot R_{BERT}}{P_{BERT} + R_{BERT}} \tag{B.16}$$

## Appendix C. Experimental configurations, counterfactual policy design, and qualitative case studies

### C.1. Counterfactual perturbation scenarios and prescriptive action modulation protocol

Counterfactual preference pairs were curated in collaboration with aquaculture domain experts to prevent the policy model from relying on language shortcuts or unconstrained visual motion. The counterfactual protocol holds underlying visual kinematics $v$ invariant while shifting environmental telemetry into biologically stressful or restrictive intervals (Table C.1). Under critical hypoxia or extreme thermal stress, restrictive actuation commands are mandated regardless of schooling vigor. This protocol enforces conditional reliance on environmental safety bounds when visual feeding activity conflicts with water quality constraints.

**Table C.1**

Counterfactual environmental perturbation scenarios and prescriptive action modulation effects in RAS.

| Environmental Dimension | Perturbation Scenarios | Physical Grounding & Biological Context | Prescribed Action Modulation Effect (Counterfactual Policy) |
|---|---|---|---|
| Water Temperature | Extreme Low (1°C) | Below critical physiological tolerance for O. mykiss | Force Stop: Feeder state, Stop; Feed amount, 0%; increase water temperature to 10-20°C |
| | Low (5°C) | Suboptimal cold stress, depressed digestive enzyme activity | Restrict Feeding: Reduce feed amount by 50%; Extend interval to 10s; increase water temperature to 10-20°C |
| | Ideal (15°C) | Optimal metabolic range and feed conversion window | Nominal Baseline: Execute action strictly according to visual kinematics; maintain current water temperature |
| | High (25°C) | Suboptimal heat stress, elevated postprandial oxygen demand | Restrict Feeding: Reduce feed amount by 20%; Extend interval to 7s; reduce water temperature to 10-20°C |
| | Extreme High (30°C) | Approaching upper lethal thermal limit | Force Stop: Feeder state, stop; Feed amount, 0%; reduce the water temperature to 10-20°C |
| Dissolved Oxygen (DO) | Critical Hypoxia (2.0 mg/L) | Below critical respiratory limit; acute asphyxia risk | Force Stop: Immediate feeding suspension to prevent DO collapse; increase DO to 8.0 mg/L |
| | Mild Hypoxia (4.0 mg/L) | Suboptimal respiratory stress | Restrict Feeding: Reduce feed amount by 50%; Extend interval to 10s; increase DO to 8.0 mg/L |
| | Hyperoxia (14.0 mg/L) | In a supersaturated state, fish may suffer from Gas Bubble Disease | Nominal Baseline: Execute action strictly according to visual kinematics; decrease DO above 8.0 mg/L |
| Stocking Density | Low Density (10 $kg/m^3$) | Minimal competition; abundant spatial resources | Nominal Baseline: Execute action strictly according to visual kinematics; moderately increase the number of fish |
| | High Density (100 $kg/m^3$) | Increased metabolic waste; elevated biofilter load | Conservative Action: Reduce feed amount by 20%; Extend interval to 7s; moderately decrease the number of fish |
| | Extreme Density (130 $kg/m^3$) | Severe spatial crowding; risk of biological filter collapse | Force Stop: Feeder state, stop to safeguard water quality; moderately decrease the number of fish |
| Feed Intake | Deficient (1% body weight) | Undersatiated baseline; high nutrient assimilation capacity | Nominal Baseline: Permit full ration according to visual appetite |
| | Ideal (4% body weight) | Standard daily nutritional ration for 65g body weight | Nominal Baseline: Execute standard threshold response |
| | Excessive (8% body weight) | Satiety limit exceeded; high residual feed pollution risk | Force Stop: Withhold feed regardless of schooling kinematics; remove residual food. |
| Fish Species | *Oreochromis niloticus* (At 12°C) | Tropical species; 12°C induces acute hypothermic coma | Force Stop: Inhibit feeding due to critical physiological cold stress |
| | *Oreochromis niloticus* (At 25°C, 8.0 mg/L DO) | Optimal tropical metabolic environment | Nominal Baseline: Permit feeding governed by visual kinematics |
| | *Channa argus* (At 12°C) | Low winter metabolic dormancy; depressed nutrient uptake | Force Stop: Inhibit feeding due to seasonal thermal dormancy |
| | Channa argus (At 20°C, 4.0 mg/L DO) | Compatible physiological baseline for the species | Nominal Baseline: Permit feeding governed by visual kinematics |

### C.2. Hardware specifications and multi-stage hyperparameter configurations

The physical experimentation and algorithmic pipeline were developed across a heterogeneous edge-cloud architecture, with hardware specifications and operating environments detailed in Table C.2. Table C.3 documents the multi-stage hyperparameter configurations across Stage 1 (HBE pretraining), Stage 2 (LoRA supervised fine-tuning), and Stage 3 (mDPO alignment).

**Table C.2**
Hardware and software experimental environment configurations.

| Platform | Component | Specification |
|---|---|---|
| Edge Computing Device | CPU | 13th Gen Intel® Core™ i9-13900K @ 3.00 GHz |
| | RAM | 64 GB DRAM |
| | GPU | NVIDIA GeForce RTX 4080 (16 GB VRAM) |
| Cloud Server | CPU | Intel® Xeon® Platinum 8352V |
| | RAM | 128 GB DRAM |
| | GPU | NVIDIA Tesla A800 (80 GB VRAM) |
| Software Stack | Operating System | Ubuntu 22.04 LTS |
| | Programming Language | Python 3.9.18 |
| | Deep Learning Framework | PyTorch 1.12.1 with CUDA 11.3 |
| | Development and Tooling | PyCharm IDE, DarkLabel |

**Table C.3**

Comprehensive hyperparameter settings and configuration details across the three training stages.

| Parameter | HBE Pretraining (Stage 1) | LLM SFT Training (Stage 2) | mDPO Training (Stage 3) |
|---|---|---|---|
| Learning Rate | $1\times10^{-4}$ | $2\times10^{-4}$ (LoRA) / $1\times10^{-6}$ (HBE) | $5\times10^{-6}$ |
| Batch Size (per device) | 64 | 8 | 2 |
| Grad Accumulation Steps | — | 2 | 1 |
| Epochs | 100 | 3 | 2 |
| Optimizer | Adam | AdamW (Dual LR) | AdamW |
| LR Scheduler | Cosine Annealing LR | Constant LR (no warmup) | Cosine (warmup_ratio=0.1) |
| Max Grad Norm | — | 1 | 0.5 |
| Max Sequence Length | — | 2048 | 768 |
| Mixed Precision | AMP (default) | bf16=True, fp16=False | bf16=False, fp16=False (float32) |
| Weight Decay | $1\times10^{-5}$ | — | — |
| Loss Weights | [1.0, 1.0, 1.0, 0.5, 0.1, 0.1] | Weak class weighting (weight=2.0) | β=0.1, α=1.0 |
| LoRA Config | — | r=8, α=16, dropout=0.05 | r=8, α=16, dropout=0.05 |
| HBE Config | K=10, M=4, D=4096, Ds=64 | K=10, M=4, D=4096 (end-to-end joint training) | K=10, M=4, D=4096 (precomputed soft tokens) |
| Data Loading Workers | 4 | 4 | 4 |
| Gradient Checkpointing | — | Optional | Optional |
| Save Strategy | Best validation accuracy | Save every epoch (max 3) | Save every epoch (max 2) |

## C.3. Qualitative behavior-grounded reasoning cases

Comparative inference transcripts evaluating Supervised Fine-Tuning (LoRA) against counterfactual mDPO in a representative strong feeding scenario are illustrated in Fig. C.1. This comparison shows how preference optimization resolves operational mismatches under identical environmental telemetry and visual prefix tokens. While the baseline Supervised Fine-Tuning model displays an action mismatch by recommending an extended 7-second feeding interval alongside contradictory text concerning feeding cessation, the preference-aligned policy links high-magnitude collective dynamics with active feeding protocols and prompt 5-second actuation verified by the deterministic validator.

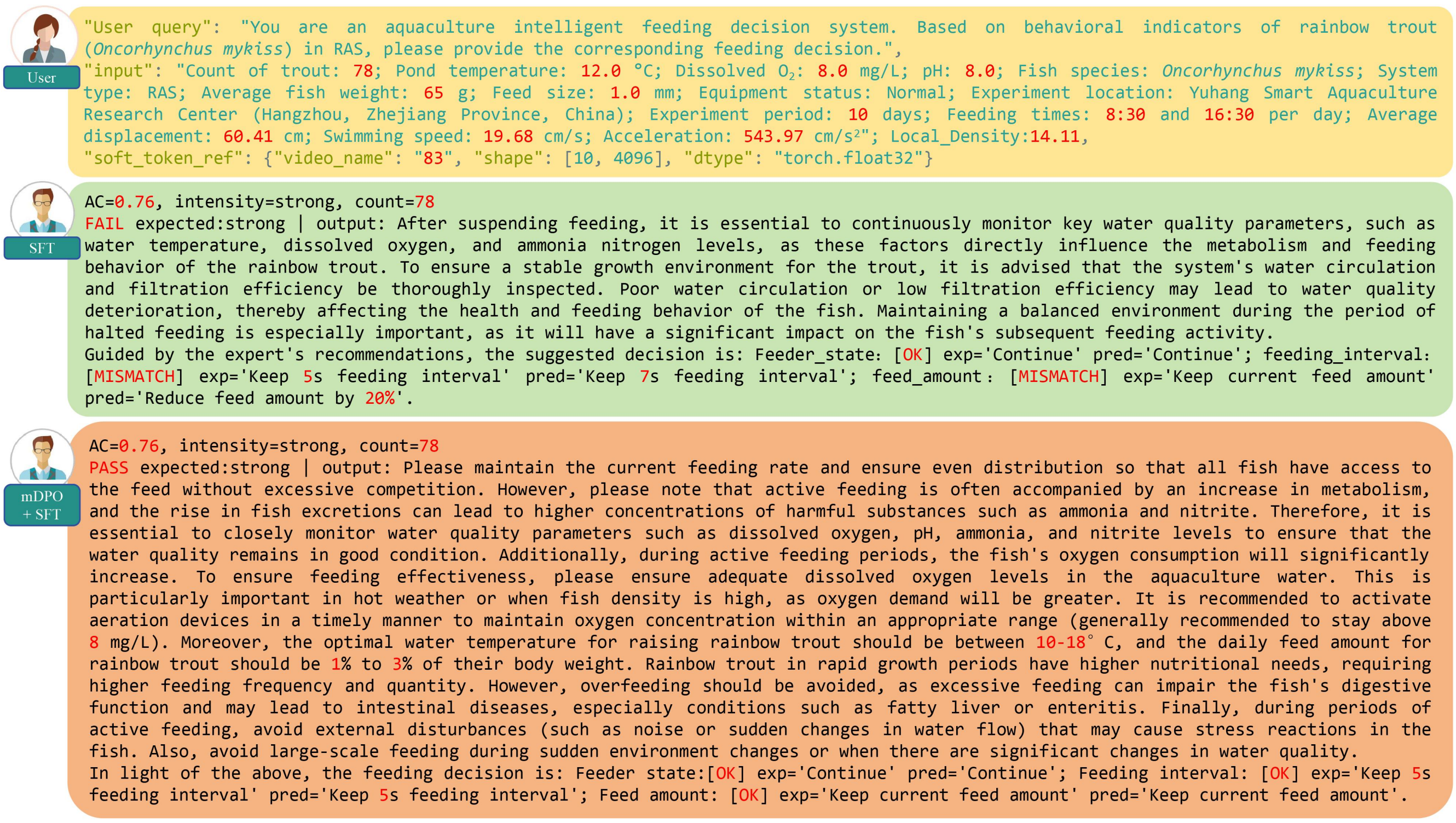

**Fig. C.1.** Qualitative comparative case study of behavior-grounded feeding decision reasoning between Supervised Fine-Tuning (LoRA) and Counterfactual mDPO.

### C.4. Quantitative operational safety and rule constraint diagnostics

The deterministic validator monitors decision outputs against expert physiological boundaries, ensuring that autonomous actuators do not induce ecological degradation or biofilter collapse in recirculating systems. As illustrated in Fig. C.2(a), the baseline Supervised Fine-Tuning (LoRA) policy yielded 28 correct decisions and 2 errors across the thirty test samples (93.33% accuracy), with one error violating restrictive feeding boundaries to trigger an unsafe continuation (USCR of 3.33%). In contrast, counterfactual mDPO achieved 29 correct decisions and 1 non-matching error (96.67% accuracy), while eliminating hazardous commands (USCR of 0.00%). This single discrepancy under mDPO represented a conservative shift from medium to weak feeding, which constitutes a fail-safe adjustment toward resource preservation. Furthermore, Fig. C.2(b) confirms that both models produced no invalid hallucinated rule formats, maintaining a 100.00% Hallucination Penalty Score (HPS) across all feeding intensity regimes. This diagnostic confirms that preference alignment enforces strict operational boundaries, intercepting abnormal events and safeguarding water quality.

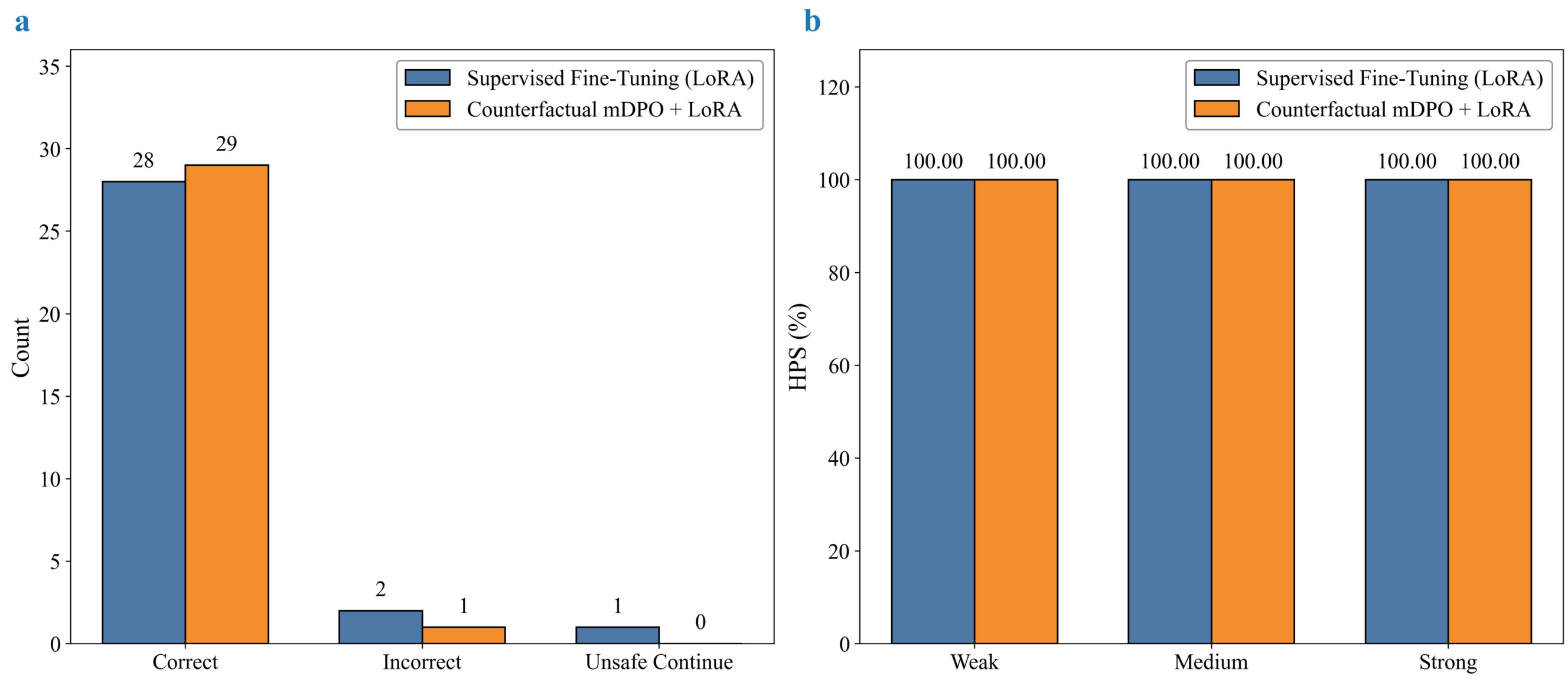

**Fig. C2.** Quantitative operational safety and constraint compliance diagnostics comparing Supervised Fine-Tuning (LoRA) and Counterfactual mDPO. (a) Frequency distributions of classification outcomes and Unsafe Continue Rate (USCR) metrics. (b) Hallucination Penalty Score (HPS) validation across feeding intensity classes.

## Appendix D. Offline full-process feeding long-video inference of THPL on juvenile *O. mykiss*

To provide an intuitive demonstration of the continuous behavioral perception and decision-making pipeline, a complete recording of the offline long-video inference of THPL is provided. Supplementary video 1 (duration: 6 min 5 s). Demonstrates the offline long-video inference and contextual decision reasoning of THPL across a complete juvenile rainbow trout (*O. mykiss*) feeding session in RAS. Operating at 5-second intervals, the system extracts continuous multi-target trajectories, computes the dynamic Activity Coefficient (AC), and synthesizes water quality constraints to generate structured feeding directives (feeder state, interval, and feed amount) along with interpretable diagnostic explanations. The demonstration video is publicly accessible at https://youtu.be/Tg2Qk7m46-A.